\documentclass{article}

    \PassOptionsToPackage{numbers, compress}{natbib}

\usepackage{enumitem}
\usepackage{wrapfig}
\usepackage{subcaption}
\usepackage{array}
\usepackage{amsmath}
\usepackage{fancyvrb}
\usepackage{fvextra}
\usepackage[final]{neurips_2026}
    \makeatletter
    \renewcommand{\@noticestring}{Workshop on Who Verifies the Agents? Toward Reliable Agent Development, NeurIPS 2026.}
    \makeatother

\usepackage[utf8]{inputenc} 
\usepackage[T1]{fontenc}    
\usepackage{hyperref}       
\usepackage{url}            
\usepackage{booktabs}       
\usepackage{amsfonts}       
\usepackage{nicefrac}       
\usepackage{microtype}      
\usepackage{xcolor}         
\usepackage{graphicx}
\usepackage{longtable}
\title{Anchor: Preventing Artifact Drift for Reliable Agent Verification}

\author{%
  Abhijay Rana
  \thanks{Equal Contribution} \\
  University of California, Berkeley\\
  \texttt{abhijayrana@berkeley.edu} \\
  \And
  Maksim Ivanov \footnotemark[1] \\
  Agentic Labs \\
  \texttt{maksim@agenticlabs.com} \\
}

\begin{document}

\maketitle

\begin{abstract}
  Reliable evaluation of long-horizon agents requires verifiers that match what the task actually asks. Verifiable rewards have driven progress in domains like math and coding, yet building similarly reliable environments for open-ended enterprise workflows repeatedly struggles to balance realism, verifiability, and scale. This creates a validity problem we call \textit{artifact drift}: when instructions, environments, oracles, and verifiers are created separately, they frequently fail to agree on what a task requires, resulting in environments that are unsolvable, reward-hackable, or inconsistent. We introduce \textbf{Anchor}, a benchmark-construction pipeline that formalizes domain expert specifications of business workflows into constrained optimization programs. From a single parametric specification, the pipeline jointly produces a natural-language instruction, environment configuration, solver-certified ground-truth solution, and state-based verifier. With this pipeline, altering parameters yields new tasks with controlled difficulty and known optimal solutions, producing harness-agnostic environments whose rewards depend solely on end-state business correctness. We apply Anchor to produce \textbf{ERP-Bench}: a benchmark of 300 long-horizon tasks spanning procurement and manufacturing workflows in a production-grade ERP system. Our pipeline organizes tasks into diverse families with calibrated difficulty profiles, and allows generated environments to evaluate coding, browser, and computer-use agents using a shared verifier. Across these harnesses, we find that frontier models follow all rules and instructions in \textbf{25.9\%} of trials and reach fully optimal decisions in only \textbf{17.4\%} of trials. Overall, Anchor and ERP-Bench offer a concrete recipe for building auditable evaluation environments whose verifiers remain aligned with the economically valuable tasks they are intended to score.
\end{abstract}

\section{Introduction}

Reliable evaluation of long-horizon agents depends on verifiers that correctly capture what a task requires. This is especially difficult for economically valuable enterprise workflows, where agents must operate over persistent business state, follow procedural rules, and make quantitatively correct decisions \citep{patwardhan2025,handa2025}. Yet even as agents improve rapidly on "real-world" benchmarks, practical deployments continue to underperform benchmark expectations \citep{pan2025,mehta2025,yehudai2025}. We argue that part of this gap comes from how agent benchmarks are constructed: the task presented to the agent and the verifier used to score it are often produced through separate processes.

A benchmark task for interactive agents consists of four artifacts: a natural-language \textit{instruction} describing the task, an \textit{environment} configuration specifying the initial state of the software the agent operates in, an \textit{oracle} solution representing a successful completion, and a \textit{verifier} that inspects the post-execution state to determine success. These artifacts must agree on what constitutes successful completion. In practice, they often do not. Most benchmarks either simplify the underlying workflow or rely on post-hoc audits to catch inconsistencies between task artifacts \citep{dsouza2026,yehudai2025}. The consequences can be severe: do-nothing agents pass 38\% of $\tau$-bench airline tasks because the verifier accepts empty responses \citep{zhu2025}. A benchmark can therefore appear realistic while still rewarding states that do not satisfy the task, rejecting valid solutions, or exposing opportunities for reward hacking.

We call this failure mode \textit{artifact drift}: the instruction, environment, oracle, and verifier describe subtly different versions of the same task. Drift is not just a verifier bug. It is a construction problem. When the four artifacts are authored independently, consistency is checked only after the fact, usually through audit and hand-written tests. These checks can catch individual mistakes, but they do not prevent systematic disagreement between what the task asks for and what the verifier rewards. At the same time, making verification easy can push benchmark designers toward overly constrained tasks with a single expected solution path \citep{zhu2025,kapoor2024}, trading realism for tractability.

We address this with \textbf{Anchor}, a benchmark-construction pipeline that derives the task and its verifier from the same solved specification. A domain expert, working with an engineer, formalizes a business workflow as a parametric constraint program in OR-Tools CP-SAT \citep{perron2025} with decision variables, business-rule constraints, and an objective metric. For a proposed parameter setting, the solver either rejects the sample as infeasible or certifies an optimal solution. The instruction, environment configuration, oracle solution, and verifier are then generated deterministically from that solved specification. Because the verifier and the task are projections of the same source, Anchor mitigates task-verifier misalignment by construction rather than relying on post-hoc repair. The same parameters also control task difficulty, allowing many task instances to be generated while preserving a consistent definition of success.

We instantiate Anchor as \textbf{ERP-Bench}: 300 tasks across procurement and manufacturing workflows in Odoo, a widely deployed open-source ERP system. Tasks span calibrated difficulty profiles and diverse workflow families within each domain. We evaluate frontier models on these tasks under three widely deployed agent paradigms: coding, browser, and computer-use. Our evaluations surface two empirical findings central to the contribution: frontier models consistently break explicit constraints rather than merely make worse tradeoffs at higher task difficulties; and UI-based agents are several times more expensive on the same task instances.

Our contributions are as follows:
\begin{itemize}[itemsep=4pt, parsep=0pt, topsep=2pt, partopsep=0pt]
    \item We introduce \textbf{Anchor}, which generates the instruction, environment, oracle, and verifier from one constrained-optimization specification.
    \item We show that single-source construction reduces cross-artifact inconsistency while preserving the need for independent specification validation.
    \item We release \textbf{ERP-Bench}, 300 long-horizon Odoo tasks spanning 29 procurement and manufacturing workflow patterns.
    \item We evaluate frontier models across coding, browser, and computer-use harnesses, finding 25.9\% constraint satisfaction but only 17.4\% fully optimal solutions.
\end{itemize}

\section{Related Work}
\label{sec:related}

\subsection{Real-world enterprise agent evaluation}

Several recent benchmarks target enterprise agent evaluation, but each addresses different parts of the realism-consistency-scale tradeoff. Expert-authored or curated tasks provide realism but are expensive to scale: SWE-bench \citep{jimenez2024} sources tasks from real GitHub issue-fix pairs through a curated pipeline, while $\tau$-bench \citep{yao2024} hand-crafts airline and retail domains. Templated environments invert this tradeoff: WebArena \citep{zhou2024} and OSWorld \citep{xie2024} scale through reusable environments atop real software, but their tasks tend toward shorter horizons and lower procedural complexity than enterprise workflows require. TheAgentCompany \citep{xu2025} pushes realism further by simulating an entire company environment, but required roughly 3,000 person-hours to construct 175 tasks.

Recent enterprise benchmarks explore other parts of the design space. EntWorld \citep{mo2026} generates 1,756 GUI tasks across CRM, ITIL, and ERP domains by reverse-engineering business logic from database schemas, though its tasks are predominantly information-retrieval queries rather than constrained operational planning. WoW-bench \citep{gupta2026} evaluates 234 ServiceNow tasks focused on cascading side-effect prediction, while Agent-Diff \citep{pysklo2026} grades 224 enterprise API tasks through state-diff contracts. None of these pipelines derive the instruction, environment, oracle, and verifier from a single solved specification, or support controlled cross-modality evaluation on identical task instances.

\subsection{Verifier reliability and artifact drift}

Verification failures are common in widely used agent benchmarks. Weak unit tests in SWE-bench can accept semantically incorrect patches, and strengthening those tests reshuffles leaderboard rankings in 40.9\% of SWE-bench Lite cases and 24.4\% of SWE-bench Verified cases \citep{aleithan2024,yu2025}. State-based verifiers can also accept unintended end states when the ground-truth state is underspecified \citep{cuadron2025}. The $\tau^2$-bench authors note that their predecessor relied on ``carefully crafted natural language instructions that help ensure a single, solvable path for the constraint satisfaction problem'' \citep{barres2025}, illustrating the tension between open-ended tasks and tractable verification.

These failures have led to ``Verified'' reissues of benchmarks such as SWE-bench and OSWorld after defects were discovered post-release \citep{openai2024swebench,xie2025osworld}. Existing responses largely operate after task construction, through curation, adversarial review, or stronger tests \citep{zhu2025,kapoor2024}. Even scalable generation pipelines such as AgentSynth \citep{xie2025agentsynth} produce task artifacts through separate generation stages, leaving consistency to downstream validation. Anchor instead moves this consistency requirement into construction itself by deriving the instruction, environment, oracle, and verifier from one solved specification.
\section{Anchor: Single-Source Task Creation}
\label{sec:anchor}

\subsection{Single-source compilation}

Many enterprise workflows operate on structured data in systems of record, follow explicit rules and actions, and optimize measurable outcomes, which makes them naturally expressible as constraint-satisfaction optimization problems. Deciding which deals a rep should work this week to hit quota, which invoices to chase first to minimize days-sales-outstanding, and which support tickets to escalate to protect SLA attainment are all problems of this form.

\begin{figure}[t]
    \centering
    \includegraphics[width=\textwidth]{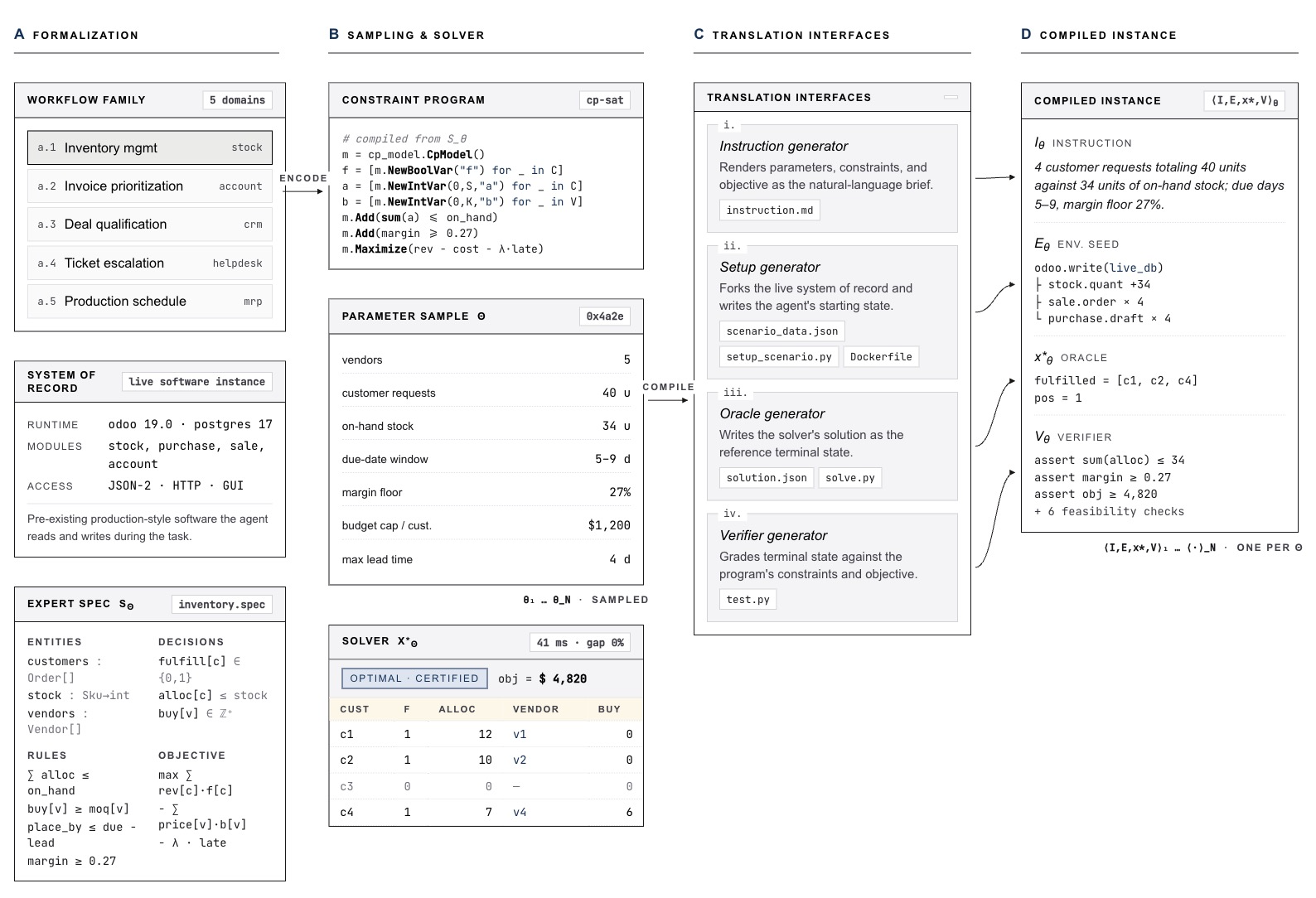}
    \caption{Anchor task generation pipeline. A workflow specification is formalized as a CP-SAT constraint program and solved under sampled parameter settings. The same solved specification generates the instruction, environment initialization, oracle solution, and verifier, while QA routes detected defects back to the source before regeneration.}
    \label{fig:pipeline}
\end{figure}

\paragraph{Pipeline overview.}
The pipeline (Figure~\ref{fig:pipeline}) turns an informal specification of a business workflow into many task and environment instances. It begins by selecting a workflow family and the system of record in which it takes place, for example invoice prioritization in an accounting system, deal qualification in a CRM, or production scheduling in an ERP. A domain expert then describes the entities involved, the decisions an agent must make, the rules governing those decisions, the economic objective, and the schema and actions exposed by the system of record. We formalize this specification as a constraint program in OR-Tools CP-SAT with decision variables, business-rule constraints, and an objective function, chosen for speed, expressive constraint vocabulary, and ability to certify optimality.

\paragraph{Single-source compilation.}
Given a parameter setting $\theta$ and a constraint program, the solver either rejects the sampled instance as infeasible or certifies an optimal solution $x^*_\theta$. We denote the resulting task instance by $(I_\theta, E_\theta, x^*_\theta, V_\theta)$, where $I_\theta$ is the natural-language instruction, $E_\theta$ is the initialized environment, $x^*_\theta$ is the solver-certified solution, and $V_\theta$ is the terminal-state verifier. All four artifacts are compiled from the same parameterized specification $S_\theta$. This means the task semantics and verifier semantics share the same source, rather than being authored independently and reconciled after the fact.

\paragraph{Translation interfaces.}
Four translation interfaces bridge the constraint program and the environment artifacts. The \textbf{instruction generator} renders parameters, constraints, and objectives as the natural-language instruction. The \textbf{setup generator} produces environment configuration and a script that seeds the system of record with the agent's starting state. The \textbf{oracle generator} writes the solver's solution into the environment as the reference terminal state. The \textbf{verifier generator} grades the terminal state against the same program constraints and objective used to define the task.

\subsection{Verifier validation and residual defects}\label{sec:residual}

Single-source generation does not eliminate all possible task construction errors. The constraint program can encode incorrect business logic, and the renderers can mistranslate a correct specification into the artifacts. It does, however, remove the need to define task semantics and verifier semantics independently for each task instance. Residual errors therefore come from the shared specification or its translation layers rather than from separately authored task artifacts drifting apart.

\paragraph{End-to-end checks.}
We use four end-to-end checks to test whether the generated task and verifier remain aligned.

\begin{itemize}[itemsep=2pt, parsep=0pt, topsep=0pt]
    \item \textbf{No-op agent.} Doing nothing should score zero. A nonzero reward would mean the initialized environment already satisfies the verifier. The no-op agent scores zero on 300/300 ERP-Bench tasks.
    \item \textbf{Oracle replay.} Executing the solver-certified solution should receive full reward. A lower score would mean the setup, oracle, and verifier disagree. Oracle replay receives full reward on 300/300 tasks.
    \item \textbf{LLM-judge consistency check.} An LLM cross-reads the instruction, environment configuration, oracle, verifier, and CP-SAT program and flags cases where an artifact disagrees with the shared specification or with the other artifacts.
    \item \textbf{Reward-hacking canary.} We flag any rollout that passes all task constraints while reporting an objective strictly better than the solver-certified optimum. Such a result would indicate either a formalization error or a hole in the verifier.
\end{itemize}

Together, these checks test complementary failure modes: false-positive rewards, false-negative rewards, cross-artifact inconsistency, and apparent violations of the certified optimum.

\paragraph{Source-level repair.}
Because the generator is a single program rather than a corpus of hand-authored tasks, any defect surfaced by these checks can be traced to the specification or one of its translation layers, fixed once, and propagated deterministically to all affected instances.

\section{ERP-Bench}\label{sec:erpbench}

We instantiate the framework as ERP-Bench, a benchmark of 300 customer order fulfillment, inventory management, and manufacturing tasks spanning 29 workflow patterns in a production-grade ERP system.

We chose the domain for its real-world value. Manufacturing contributes trillions of dollars in annual U.S.\ value added, accounts for roughly a tenth of U.S.\ GDP, and employs roughly 12.6M workers across more than 239,000 firms \citep{bea2025,nam2025}. Procurement is a core horizontal function, with roughly 58,700 annual openings projected for purchasing managers, buyers, and purchasing agents through 2034 \citep{bls2025}.

\subsection{Environment}\label{sec:environment}

Tasks run in Odoo 19, a widely deployed open-source ERP, packaged one container per task with a fresh database seeded with inventory, customers, vendors, bills of materials, and work center data. Agents operate against the same persistent records a back-office user would touch --- sales orders, purchase orders, manufacturing orders, vendor pricelists, and invoices --- through the JSON-2 API or the standard Odoo web client. We make use of Odoo's sales, inventory, manufacturing, and invoicing modules; its highly modular architecture spans more than 50 first-party apps on a shared schema, so our task creation methodology can be extended to a wide range of business operations beyond this release.

\subsection{Task Design}\label{sec:task-design}

We worked with ten freelance ERP practitioners with 5+ years of experience implementing or operating production ERP systems to ground the workflow families in actual back-office practice. The collaboration spanned roughly twenty hours of elicitation interviews and another twenty hours of task review once early instances were generated. This 40-person-hour investment produced 29 reusable workflow patterns, which the pipeline sampled into 300 task instances; the marginal cost per additional task instance approaches zero, since expert effort is incurred per family rather than per task.

The interviews produced four kinds of input that inform our CP-SAT model:

\begin{itemize}[itemsep=2pt, parsep=0pt, topsep=0pt]
    \item \textbf{Industry domains} such as furniture wholesalers, electrical equipment manufacturers, and lab equipment dealers.
    \item \textbf{Entities} such as customers, vendors, products, work centers, and bills of materials.
    \item \textbf{Constraints} around suppliers and offers, lead times, invoicing policies, inventory and order quantities, and ERP hygiene rules like record linkage and provenance.
    \item \textbf{Objectives} dominated by minimizing economic spending, with secondary preferences for using fewer vendors, consolidating orders, and preserving free manufacturing capacity.
\end{itemize}

The translation layer specifies requests and objectives in the natural-language instruction, while constraints and business rules are dispersed both in the instruction and the ERP state itself, mirroring real-world workflow messiness.

\subsection{Verification and Reward Shaping}\label{sec:verification}

ERP-Bench tasks are designed to remain open-ended while still supporting deterministic verification. The CP-SAT solver certifies an exact optimal objective value, but many terminal states can achieve that optimum. For example, two customers may each order 10 units of the same product, while the warehouse holds 5 units and two suppliers offer the remaining units at the same price and lead time. The solver certifies the minimum spend, but the agent can choose either supplier, allocate existing stock to either customer, and place orders on any schedule that still satisfies both deadlines. The verifier accepts any terminal state that satisfies the task constraints and matches the certified optimum.

Verifiers grade three dimensions: constraint satisfaction, optimality, and traceability. These combine into a 0--100 reward weighted 25\% constraints, 60\% optimality, and 15\% traceability. Constraint satisfaction gates the other dimensions: if any explicit task constraint fails, the agent receives only the partial constraint score, and optimality and traceability credit are removed. A small set of structural violations act as hard zeros.

The verifier evaluates the final ERP state rather than requiring a particular action sequence. Constraint checks cover demand fulfillment, deadlines, supplier rules, inventory limits, manufacturing feasibility, invoicing, and other task-specific requirements. Optimality is computed from the realized terminal state and compared against the solver-certified objective. Traceability checks that the records created by the agent remain linked to the business transactions they serve.

\paragraph{Verifier validation.}
We validate the generated verifiers against both positive and negative controls. Oracle replay receives full credit on all 300 tasks, while a no-op agent receives zero on all 300. Verifier complexity also scales with task difficulty: easy tasks instantiate 55.76 checks on average, compared with 73.01 for medium tasks and 122.76 for hard tasks.

\subsection{Difficulty and Diversity Engineering}\label{sec:difficulty}

We compose configuration parameters along several independent axes, then sample customer rosters, vendor rosters, order quantities, deadlines, stock levels, prices, lead times, and capacity limits from ranges under a fixed seed, rotating through industrial product domains and seeding adjacent records to mirror production ERP state. The same parameters control difficulty, scaling demand and tightening supply while progressively unlocking manufacturing surface and enriching the objective from feasibility to multi-criterion optimization. The result is a dataset whose tasks differ in structure, scale, objective, surface language, and optimal plan, while remaining verifiable.

\subsection{Generation}\label{sec:generation}

To produce 300 tasks, we sample parameter settings uniformly from configured ranges (Appendix~\ref{app:ranges}) and submit each to the CP-SAT solver. The full pipeline runs in approximately 11 minutes of wall time, with Anchor rejecting 732 samples as infeasible: 275 by pre-solver heuristics and 457 by solver timeout (retried at 15 seconds before final rejection). For successful solves, an additional lexicographic tie-breaker fixes the selected solution so that regeneration runs produce identical artifacts.

\begin{table}[t]
    \centering
    \small
    \caption{Solver performance and model size by difficulty tier. Solve-time columns report cumulative CP-SAT wall time per accepted task, including failed solver calls and retries. Solve time scales over 100$\times$ from easy to hard tasks while decision variable counts grow roughly 5$\times$.}
    \label{tab:difficulty-breakdown}
\begin{tabular}{lrrrrrrrr}
    \toprule
    Difficulty & Tasks & Avg Solve & Median & P95 & Max & Avg Decision Vars & Discards & Resampled \\
    \midrule
    Easy & 50 & 0.044s & 0.008s & 0.288s & 0.347s & 19.8 & 253 & 20 \\
    Medium & 166 & 0.951s & 0.031s & 2.842s & 20.863s & 78.6 & 243 & 79 \\
    Hard & 84 & 5.856s & 0.100s & 31.984s & 100.735s & 96.7 & 236 & 56 \\
    \bottomrule
\end{tabular}
\end{table}


\paragraph{Dataset} The release contains 300 tasks across 29 diverse business workflow patterns, with per-pattern and per-family counts shown in Appendix~\ref{app:dataset}. Each task includes a generator config and seed for reproducibility. We release the data in Harbor task format, with separate instruction, task configuration, environment, solution, and verifier files \url{https://huggingface.co/datasets/abhijayrana/erp-bench}.

\section{Evaluation}\label{sec:eval}

We evaluate five frontier models on ERP-Bench across three agent harnesses: coding, browser, and computer-use. Each harness operates on the same task instances and is graded by the same verifier. Across 12 model-harness combinations, we conduct 16,159 agent trials, with two computer-use configurations stopped early.

\subsection{Agents}\label{sec:agents}

We build our agent scaffolds on the open-source pi-mono agent toolkit \citep{zechner2026}, with browser and computer-use modes added as extensions. The coding harness uses the default pi agent configuration with shell and filesystem tools, driving Odoo through the JSON-2 API. The browser harness extends pi with a Playwright-based tool that drives the standard Odoo web client through a11y tree-resolved actions, with access only to that tool. The computer-use harness drives an Xvfb-backed Chromium through pixel-coordinate clicks, keystrokes, and screenshots, with no DOM access and no coding tools. Implementation details are in Appendix~\ref{app:harness}.

The coding agent scaffold was chosen because recent coding agents show strong results on terminal and broader computer tasks \citep{merrill2026,openai2026,anthropic2026}; browser and computer-use scaffolds were evaluated because they resemble the workflow a human would use. All three point at the same per-task containerized environment and are graded by the same verifier. We run 5 trials per agent-task pair with no retries, a 1-hour per-trial timeout, 400-turn budgets, and reasoning effort set to each model's provider default where exposed. Computer-use runs for open-weight models were halted early after over 500 trials produced zero points; those cells reflect partial trials.

\subsection{Models}\label{sec:models}

We select top proprietary models (GPT-5.5, Claude Opus 4.7) and top open-weight models (GLM-5.1, GLM-5v, Kimi K-2.5). We evaluate each model with every harness, with one exception: GLM-5.1 does not natively support vision inputs and is incompatible with the computer-use harness, so we replace it with the closest vision-capable equivalent, GLM-5V-Turbo. Models with configurable reasoning effort are run with the provider's default (i.e.\ ``medium'' for Anthropic and OpenAI models). Proprietary models are accessed via first-party developer APIs, open-weight models via OpenRouter.

\begin{figure}
    \centering
    \includegraphics[height=0.35\textheight]{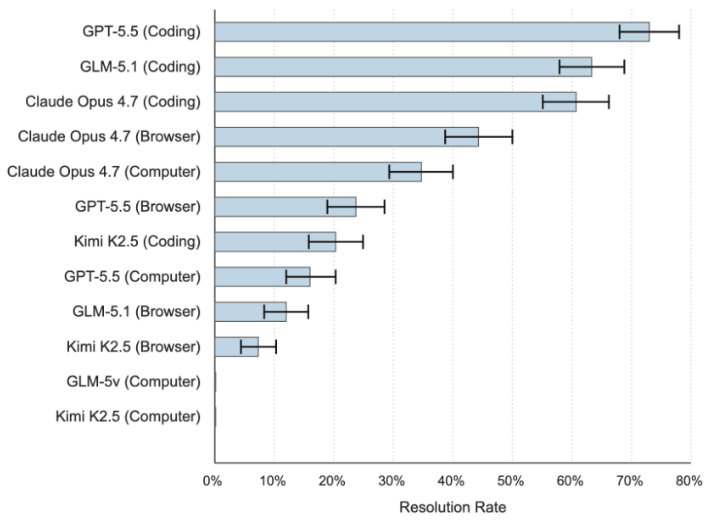}
    \caption{Resolution rates by model and harness. Bars report resolution rates (pass@5) over the 300-task release for each evaluated model-harness combination, with error bars showing uncertainty across repeated runs.}
    \label{fig:resolution-rates}
\end{figure}

\paragraph{Verifier robustness.}
Across all 16,159 agent trials, no rollout reached a constraint-feasible terminal state with an objective strictly better than the solver-certified optimum. Thus, no evaluated agent found a terminal state that satisfied the verifier's constraints while exceeding the formal optimum.

\subsection{Generated Difficulty Tracks Realized Agent Difficulty}\label{sec:difficulty-correlation}

The generator's parameters translate into harder tasks for every model in every harness. Across the 300-task set, the strongest negative Spearman correlations against task-level pass@1 come from scale and structure variables: workcenters, manufacturing orders, solver variables, solver constraints, BOM component lines, purchase orders, and distinct vendors. Maximum BOM depth and on-hand-stock-to-demand ratio sit near zero because the sampler clamps depth tightly and equalizes stock pressure across difficulty buckets. The declared easy/medium/hard bands collapse to the same monotone signal in every harness, with aggregate pass@5 falling from 70.5\% on easy to 22.3\% on hard for coding agents, from 46.5\% to 7.7\% for browser agents, and from 56.0\% to 9.5\% for computer-use agents. The single positive correlate, pre-seeded sales orders (+0.107), reflects that more existing orders means less work for the agent.


For resolved trials, task parameters also correlate with the number of steps the agent takes to resolve a task. We break out correlations by harness since harnesses are not directly comparable in step count.


\paragraph{Feasibility-optimality gap.}
Headline aggregate percentages are computed over the 15,000 trials from ten complete model-harness configurations, excluding the two early-stopped computer-use configurations. Across all interfaces, agents satisfy every hard operational constraint in 25.9\% of trials but reach the certified optimum in only 17.4\%. Among constraint-clean trials, mean optimality is 93.4, so agents that remain feasible usually make near-optimal decisions. Performance degradation at higher difficulty therefore comes primarily from violating explicit task constraints rather than making poor tradeoffs within the feasible region.
\begin{figure}
    \centering
    \includegraphics[height=0.35\textheight]{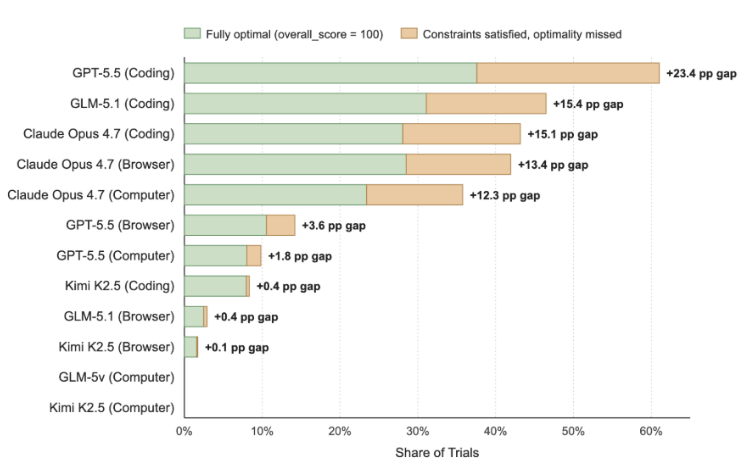}
    \caption{Feasibility and optimality split by model and harness. Green bars show trials with full reward, and tan extensions show trials that satisfy all hard constraints without reaching the certified optimum.}
    \label{fig:feasibility-optimality}
\end{figure}

\begin{figure}
    \centering
    \includegraphics[height=0.35\textheight]{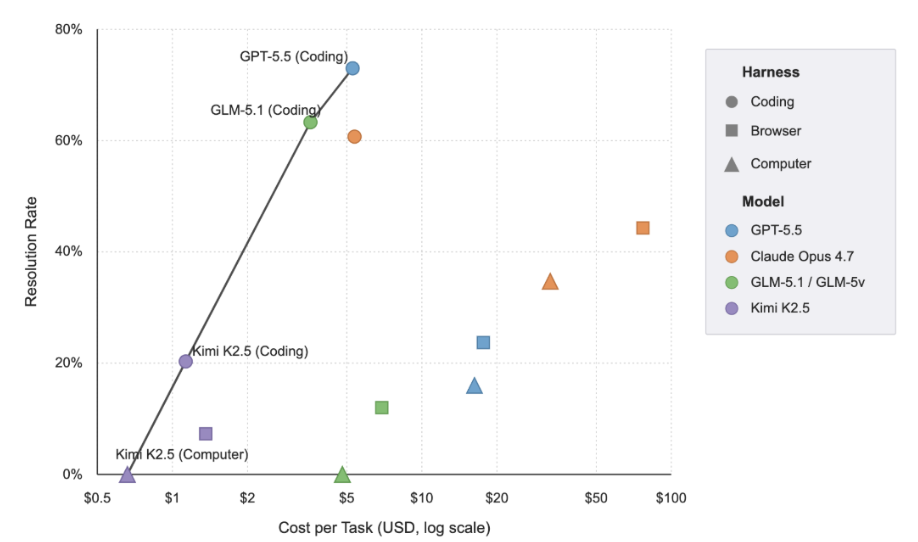}
    \caption{Cost and resolution by model and harness. The coding harness dominates the cost–quality frontier across all four evaluated models. The black line traces the cost–quality frontier across coding agents.}
    \label{fig:cost-resolution}
\end{figure}

\paragraph{Agent harness comparisons.}
Despite ERP-Bench not being a coding benchmark, coding agents are both more effective and cheaper than GUI-based agents on identical task instances. Claude Opus 4.7 achieves 60.7\%, 44.3\%, and 38.3\% pass@5 across coding, browser, and computer-use respectively. Model rankings also change across interfaces: GPT-5.5 leads Claude in coding, while Claude leads in both GUI harnesses. Cost follows the same pattern (Figure~\ref{fig:cost-resolution}): GPT-5.5 costs \$5.28 per task in coding versus \$17.65 in browser and \$16.25 in computer-use, while Claude costs \$5.37, \$76.88, and \$32.71 respectively. Two of four models scored zero on every completed computer-use trial. Because the harnesses differ in tool access and interaction mechanisms, these results compare end-to-end agent systems rather than isolating the causal effect of interface modality.

\section{Limitations}\label{sec:limitations}

The main cost of our task construction is paid upfront in formalization: a domain expert specifies the workflow, an engineer writes a CP-SAT program modelling the entities, constraints, objective, and ERP actions, and a translation layer keeps artifacts aligned. Task validity still depends on the specification. Single-source generation reduces disagreement between artifacts, but it does not guarantee that the shared specification itself is correct. A verifier can therefore remain perfectly aligned with an instruction and oracle while still encoding an incomplete or mistaken model of the underlying business process. A constraint program can omit a business rule, a renderer can express the right rule unclearly, and an ERP can expose defaults the formal model did not intend to make relevant. The benchmark covers the part of enterprise work that is naturally visible in terminal system state. Workflows depending on tacit managerial judgment, negotiation, free-text persuasion, or outcomes outside the system of record would not fit well to this framework. Generated instructions are also more explicit than many real workplace requests. ERP-Bench prioritizes feasible, reproducible, and exactly gradable tasks, so released instructions expose the constraints and objectives needed for a fair terminal-state reward.

\section{Conclusion}\label{sec:conclusion}

We introduce Anchor, a single-source benchmark-construction pipeline that compiles instructions, environments, oracle solutions, and verifiers from one solved specification, and ERP-Bench, a benchmark of 300 long-horizon procurement and manufacturing tasks generated with Anchor. Across three agent harnesses, frontier models satisfy hard operational constraints in 25.9\% of trials but reach the certified optimum in only 17.4\%. Across 16,159 trials, no agent reached a constraint-feasible terminal state with an objective better than the solver-certified optimum. More broadly, inconsistencies between independently authored task artifacts are a recurring source of verifier failure in agent benchmarks. Anchor addresses this by making the task and its verifier projections of the same formal specification, shifting consistency from post-hoc audit into benchmark construction.

\begin{ack}
\textbf{Funding.} This research was supported by Agentic Labs.

\end{ack}

\medskip

{
\small

\bibliographystyle{plainnat}
\bibliography{references}

@misc{patwardhan2025,
  title={GDPval: Evaluating AI Model Performance on Real-World Economically Valuable Tasks},
  author={Patwardhan, Tejal and Dias, Rachel and Proehl, Elizabeth and Kim, Grace and Wang, Michele and Watkins, Olivia and Fishman, Sim{\'o}n Posada and Aljubeh, Marwan and Thacker, Phoebe and Fauconnet, Laurance and Kim, Natalie S. and Chao, Patrick and Miserendino, Samuel and Chabot, Gildas and Li, David and Sharman, Michael and Barr, Alexandra and Glaese, Amelia and Tworek, Jerry},
  year={2025},
  eprint={2510.04374},
  archivePrefix={arXiv},
  primaryClass={cs.AI},
  url={https://arxiv.org/abs/2510.04374}
}

@misc{handa2025,
  title={Which Economic Tasks are Performed with AI? Evidence from Millions of Claude Conversations},
  author={Handa, Kunal and Tamkin, Alex and McCain, Miles and Huang, Saffron and Durmus, Esin and Heck, Sarah and Mueller, Jared and Hong, Jerry and Ritchie, Stuart and Belonax, Tim and Troy, Kevin K. and Amodei, Dario and Kaplan, Jared and Clark, Jack and Ganguli, Deep},
  year={2025},
  eprint={2503.04761},
  archivePrefix={arXiv},
  primaryClass={cs.CY},
  url={https://arxiv.org/abs/2503.04761}
}

@misc{pan2025,
  title={Measuring Agents in Production},
  author={Pan, Melissa Z. and Arabzadeh, Negar and Cogo, Riccardo and Zhu, Yuxuan and Xiong, Alexander and Agrawal, Lakshya A and Mao, Huanzhi and Shen, Emma and Pallerla, Sid and Patel, Liana and Liu, Shu and Shi, Tianneng and Liu, Xiaoyuan and Davis, Jared Quincy and Lacavalla, Emmanuele and Basile, Alessandro and Yang, Shuyi and Castro, Paul and Kang, Daniel and Gonzalez, Joseph E. and Sen, Koushik and Song, Dawn and Stoica, Ion and Zaharia, Matei and Ellis, Marquita},
  year={2025},
  eprint={2512.04123},
  archivePrefix={arXiv},
  primaryClass={cs.SE},
  url={https://arxiv.org/abs/2512.04123}
}

@misc{mehta2025,
  title={Beyond Accuracy: A Multi-Dimensional Framework for Evaluating Enterprise Agentic AI Systems},
  author={Mehta, Sushant},
  year={2025},
  eprint={2511.14136},
  archivePrefix={arXiv},
  primaryClass={cs.AI},
  url={https://arxiv.org/abs/2511.14136}
}

@misc{yehudai2025,
  title={Survey on Evaluation of LLM-based Agents},
  author={Yehudai, Asaf and Eden, Lilach and Li, Alan and Uziel, Guy and Zhao, Yilun and Bar-Haim, Roy and Cohan, Arman and Shmueli-Scheuer, Michal},
  year={2025},
  eprint={2503.16416},
  archivePrefix={arXiv},
  primaryClass={cs.AI},
  url={https://arxiv.org/abs/2503.16416}
}

@misc{dsouza2026,
  title={Benchmarking Agents in Insurance Underwriting Environments},
  author={Dsouza, Amanda and Ramakrishnan, Ramya and Dickens, Charles and Pohani, Bhavishya and Glaze, Christopher M.},
  year={2026},
  eprint={2602.00456},
  archivePrefix={arXiv},
  primaryClass={cs.AI},
  url={https://arxiv.org/abs/2602.00456}
}

@misc{zhu2025,
  title={Establishing Best Practices for Building Rigorous Agentic Benchmarks},
  author={Zhu, Yuxuan and Jin, Tengjun and Pruksachatkun, Yada and Zhang, Andy and Liu, Shu and Cui, Sasha and Kapoor, Sayash and Longpre, Shayne and Meng, Kevin and Weiss, Rebecca and Barez, Fazl and Gupta, Rahul and Dhamala, Jwala and Merizian, Jacob and Giulianelli, Mario and Coppock, Harry and Ududec, Cozmin and Sekhon, Jasjeet and Steinhardt, Jacob and Kellermann, Antony and Schwettmann, Sarah and Zaharia, Matei and Stoica, Ion and Liang, Percy and Kang, Daniel},
  year={2025},
  eprint={2507.02825},
  archivePrefix={arXiv},
  primaryClass={cs.AI},
  url={https://arxiv.org/abs/2507.02825}
}

@misc{kapoor2024,
  title={AI Agents That Matter},
  author={Kapoor, Sayash and Stroebl, Benedikt and Siegel, Zachary S. and Nadgir, Nitya and Narayanan, Arvind},
  year={2024},
  eprint={2407.01502},
  archivePrefix={arXiv},
  primaryClass={cs.LG},
  url={https://arxiv.org/abs/2407.01502}
}

@inproceedings{perron2025,
  title={The {CP-SAT-LP} Solver},
  author={Perron, Laurent and Didier, Fr{\'e}d{\'e}ric and Gay, Steven},
  booktitle={29th International Conference on Principles and Practice of Constraint Programming (CP 2023)},
  series={Leibniz International Proceedings in Informatics (LIPIcs)},
  volume={280},
  pages={3:1--3:2},
  year={2023},
  publisher={Schloss Dagstuhl -- Leibniz-Zentrum f{\"u}r Informatik},
  doi={10.4230/LIPIcs.CP.2023.3}
}

@inproceedings{jimenez2024,
  title={{SWE}-bench: Can Language Models Resolve Real-World {GitHub} Issues?},
  author={Jimenez, Carlos E. and Yang, John and Wettig, Alexander and Yao, Shunyu and Pei, Kexin and Press, Ofir and Narasimhan, Karthik R.},
  booktitle={The Twelfth International Conference on Learning Representations},
  year={2024},
  url={https://openreview.net/forum?id=VTF8yNQM66}
}

@misc{yao2024,
  title={{$\tau$}-bench: A Benchmark for Tool-Agent-User Interaction in Real-World Domains},
  author={Yao, Shunyu and Shinn, Noah and Razavi, Pedram and Narasimhan, Karthik},
  year={2024},
  eprint={2406.12045},
  archivePrefix={arXiv},
  primaryClass={cs.AI},
  url={https://arxiv.org/abs/2406.12045}
}

@inproceedings{zhou2024,
  title={{WebArena}: A Realistic Web Environment for Building Autonomous Agents},
  author={Zhou, Shuyan and Xu, Frank F. and Zhu, Hao and Zhou, Xuhui and Lo, Robert and Sridhar, Abishek and Cheng, Xianyi and Ou, Tianyue and Bisk, Yonatan and Fried, Daniel and Alon, Uri and Neubig, Graham},
  booktitle={The Twelfth International Conference on Learning Representations},
  year={2024},
  url={https://openreview.net/forum?id=oKn9c6ytLx}
}

@inproceedings{xie2024,
  title={{OSWorld}: Benchmarking Multimodal Agents for Open-Ended Tasks in Real Computer Environments},
  author={Xie, Tianbao and Zhang, Danyang and Chen, Jixuan and Li, Xiaochuan and Zhao, Siheng and Cao, Ruisheng and Hua, Toh Jing and Cheng, Zhoujun and Shin, Dongchan and Lei, Fangyu and Liu, Yitao and Xu, Yiheng and Zhou, Shuyan and Savarese, Silvio and Xiong, Caiming and Zhong, Victor and Yu, Tao},
  booktitle={Advances in Neural Information Processing Systems},
  year={2024},
  url={https://arxiv.org/abs/2404.07972}
}

@misc{xu2025,
  title={{TheAgentCompany}: Benchmarking {LLM} Agents on Consequential Real World Tasks},
  author={Xu, Frank F. and Song, Yufan and Li, Boxuan and Tang, Yuxuan and Jain, Kritanjali and Bao, Mengxue and Wang, Zora Z. and Zhou, Xuhui and Guo, Zhitong and Cao, Murong and Yang, Mingyang and Lu, Hao Yang and Martin, Amaad and Su, Zhe and Maben, Leander and Mehta, Raj and Chi, Wayne and Jang, Lawrence and Xie, Yiqing and Zhou, Shuyan and Neubig, Graham},
  year={2024},
  eprint={2412.14161},
  archivePrefix={arXiv},
  primaryClass={cs.CL},
  url={https://arxiv.org/abs/2412.14161}
}

@misc{mo2026,
  title={{EntWorld}: A Holistic Environment and Benchmark for Verifiable Enterprise {GUI} Agents},
  author={Mo, Ying and Bai, Yu and Sun, Dapeng and Shi, Yuqian and Miao, Yukai and Chen, Li and Li, Dan},
  year={2026},
  eprint={2601.17722},
  archivePrefix={arXiv},
  primaryClass={cs.AI},
  url={https://arxiv.org/abs/2601.17722}
}

@misc{gupta2026,
  title={World of Workflows: A Benchmark for Bringing World Models to Enterprise Systems},
  author={Gupta, Lakshya and Li, Litao and Liu, Yizhe and Subramanian, Sriram Ganapathi and Suleman, Kaheer and Zhang, Zichen and Lu, Haoye and Pasupalak, Sumit},
  year={2026},
  eprint={2601.22130},
  archivePrefix={arXiv},
  primaryClass={cs.AI},
  url={https://arxiv.org/abs/2601.22130}
}

@misc{pysklo2026,
  title={Agent-Diff: Benchmarking {LLM} Agents on Enterprise {API} Tasks via Code Execution with State-Diff-Based Evaluation},
  author={Pysklo, Hubert M. and Zhuravel, Artem and Watson, Patrick D.},
  year={2026},
  eprint={2602.11224},
  archivePrefix={arXiv},
  primaryClass={cs.AI},
  url={https://arxiv.org/abs/2602.11224}
}

@misc{aleithan2024,
  title={{SWE-Bench+}: Enhanced Coding Benchmark for {LLMs}},
  author={Aleithan, Reem and Xue, Haoran and Mohajer, Mohammad Mahdi and Nnorom, Elijah and Uddin, Gias and Wang, Song},
  year={2024},
  eprint={2410.06992},
  archivePrefix={arXiv},
  primaryClass={cs.SE},
  url={https://arxiv.org/abs/2410.06992}
}

@inproceedings{yu2025,
  title={{UTBoost}: Rigorous Evaluation of Coding Agents on {SWE-Bench}},
  author={Yu, Boxi and Zhu, Yuxuan and He, Pinjia and Kang, Daniel},
  booktitle={Proceedings of the 63rd Annual Meeting of the Association for Computational Linguistics (Volume 1: Long Papers)},
  pages={3762--3774},
  year={2025},
  publisher={Association for Computational Linguistics},
  url={https://aclanthology.org/2025.acl-long.189/}
}

@misc{cuadron2025,
  title={The Danger of Overthinking: Examining the Reasoning-Action Dilemma in Agentic Tasks},
  author={Cuadron, Alejandro and Li, Dacheng and Ma, Wenjie and Wang, Xingyao and Wang, Yichuan and Zhuang, Siyuan and Liu, Shu and Schroeder, Luis Gaspar and Xia, Tian and Mao, Huanzhi and Thumiger, Nicholas and Desai, Aditya and Stoica, Ion and Klimovic, Ana and Neubig, Graham and Gonzalez, Joseph E.},
  year={2025},
  eprint={2502.08235},
  archivePrefix={arXiv},
  primaryClass={cs.AI},
  url={https://arxiv.org/abs/2502.08235}
}

@misc{barres2025,
  title={{$\tau^2$}-Bench: Evaluating Conversational Agents in a Dual-Control Environment},
  author={Barres, Victor and Dong, Honghua and Ray, Soham and Si, Xujie and Narasimhan, Karthik},
  year={2025},
  eprint={2506.07982},
  archivePrefix={arXiv},
  primaryClass={cs.AI},
  url={https://arxiv.org/abs/2506.07982}
}

@misc{openai2024swebench,
  title={Introducing {SWE}-bench {Verified}},
  author={{OpenAI}},
  year={2024},
  howpublished={\url{https://openai.com/index/introducing-swe-bench-verified/}},
  note={Blog post}
}

@misc{xie2025osworld,
  title={Introducing {OSWorld}-{Verified}},
  author={Xie, Tianbao and Yuan, Mengqi and Zhang, Danyang and Xiong, Xinzhuang and Shen, Zhennan and Zhou, Zilong and Wang, Xinyuan and Chen, Yanxu and Deng, Jiaqi and Chen, Junda and Wang, Bowen and Wu, Haoyuan and Chen, Jixuan and Wang, Junli and Lu, Dunjie and Hu, Hao and Yu, Tao},
  year={2025},
  howpublished={\url{https://xlang.ai/blog/osworld-verified}},
  note={Blog post, July 2025}
}

@misc{xie2025agentsynth,
  title={{AgentSynth}: Scalable Task Generation for Generalist Computer-Use Agents},
  author={Xie, Jingxu and Xu, Dylan and Zhao, Xuandong and Song, Dawn},
  year={2025},
  eprint={2506.14205},
  archivePrefix={arXiv},
  primaryClass={cs.AI},
  url={https://arxiv.org/abs/2506.14205}
}

@misc{bea2025,
  author = {{U.S. Bureau of Economic Analysis}},
  title = {Gross Domestic Product, 4th Quarter and Year 2024 (Third Estimate), {GDP} by Industry, and Corporate Profits},
  year = {2025},
  howpublished = {\url{https://www.bea.gov/news/2025/gross-domestic-product-4th-quarter-and-year-2024-third-estimate-gdp-industry-and}},
  note = {Accessed: 2026-05-07}
}

@misc{nam2025,
  author = {{National Association of Manufacturers}},
  title = {Facts About Manufacturing},
  year = {2025},
  howpublished = {\url{https://nam.org/mfgdata/facts-about-manufacturing-expanded/}},
  note = {Accessed: 2026-05-07}
}

@misc{bls2025,
  author = {{U.S. Bureau of Labor Statistics}},
  title = {Occupational Outlook Handbook: Purchasing Managers, Buyers, and Purchasing Agents},
  year = {2025},
  howpublished = {\url{https://www.bls.gov/ooh/business-and-financial/purchasing-managers-buyers-and-purchasing-agents.htm}},
  note = {Accessed: 2026-05-07}
}

@misc{zechner2026,
  author = {Zechner, Mario},
  title = {pi-mono: A Minimal {AI} Agent Toolkit},
  year = {2026},
  howpublished = {\url{https://github.com/badlogic/pi-mono}},
  note = {Accessed: 2026-05-07}
}

@misc{merrill2026,
  title = {Terminal-{B}ench: Benchmarking Agents on Hard, Realistic Tasks in Command Line Interfaces},
  author = {Merrill, Mike A. and Shaw, Alexander G. and Carlini, Nicholas and Li, Boxuan and Raj, Harsh and Bercovich, Ivan and Shi, Lin and Shin, Jeong Yeon and Walshe, Thomas and Buchanan, E. Kelly and others},
  year = {2026},
  eprint = {2601.11868},
  archivePrefix = {arXiv},
  primaryClass = {cs.AI},
  url = {https://arxiv.org/abs/2601.11868}
}

@misc{openai2026,
  author = {{OpenAI}},
  title = {Codex {CLI}: A Lightweight Coding Agent for the Terminal},
  year = {2026},
  howpublished = {\url{https://github.com/openai/codex}},
  note = {Accessed: 2026-05-07}
}

@misc{anthropic2026,
  author = {{Anthropic}},
  title = {Claude Code: An Agentic Coding Tool for the Terminal},
  year = {2026},
  howpublished = {\url{https://www.anthropic.com/product/claude-code}},
  note = {Accessed: 2026-05-07}
}
}


\appendix
\begingroup
\small
\setlength{\tabcolsep}{4pt}
\let\OriginalUnderscore\_
\renewcommand{\_}{\OriginalUnderscore\discretionary{}{}{}}

\section{ERP Glossary}\label{app:glossary}

\subsection{General ERP Terms}\label{a.1-general-erp-terms}

\textbf{ERP.} ERP stands for enterprise resource planning. An ERP is the system of record that a firm uses to run business operations such as sales, purchasing, inventory, manufacturing, accounting, and fulfillment. ERP-Bench uses Odoo as the ERP system and grades the final state of Odoo records after an agent acts.

\textbf{System of record.} A system of record is the authoritative database for an operational process. In ERP-Bench, Odoo is the system of record, so the verifier treats the final Odoo database state as the source of truth.

\textbf{Record.} A record is one stored business object in the ERP, such as a customer, product, sales order, purchase order, invoice, or manufacturing order. ERP-Bench asks agents to create and update records that satisfy the task instructions.

\textbf{Fulfillment path.} A fulfillment path is the chain of records that covers customer demand, for example SO to PO or SO to MO to component PO.

\textbf{Lead time.} Lead time is the delay between placing an order and receiving usable supply. ERP-Bench uses lead times to decide whether a purchase or manufacturing plan can meet customer deadlines.

\textbf{Minimum and maximum order quantity.} A minimum order quantity is the smallest amount a supplier will sell under an offer. A maximum order quantity is the largest amount allowed under the task horizon or vendor policy.

\textbf{On-hand stock.} On-hand stock is inventory already available in the warehouse. Agents can allocate existing stock to customer demand before buying or manufacturing additional units.

\textbf{Margin.} Margin is the share of selling revenue not consumed by the cost of new supply, expressed as a percentage. Some ERP-Bench tasks require portfolio-level new-spend margin to clear a stated threshold.

\textbf{Adjacent data.} Adjacent data is business state present in the ERP that belongs to the same company context and is unrelated to the task. ERP-Bench checks that agents leave adjacent records untouched.

\textbf{Terminal state.} The terminal state is the final ERP database state after the agent finishes. ERP-Bench grades the terminal state without giving credit for the trajectory itself.

\subsection{Odoo-Specific Terms}\label{a.2-odoo-specific-terms}

\textbf{Odoo.} \href{https://www.odoo.com/documentation/19.0/}{Odoo} is the open-source ERP used in ERP-Bench. The benchmark uses Odoo sales, inventory, purchasing, manufacturing, and invoicing records as the persistent business state.

\textbf{Sales order.} A sales order, or SO, records a customer commitment. In ERP-Bench, sales orders specify who ordered a product, how many units they need, what price they receive, and when the order must be delivered. Odoo documents the sales workflow in the \href{https://www.odoo.com/documentation/19.0/applications/sales/sales.html}{Sales app documentation}.

\textbf{Purchase order.} A purchase order, or PO, records a supplier commitment. In ERP-Bench, agents create purchase orders when they need vendors to supply finished goods or components. Odoo documents purchasing in the \href{https://www.odoo.com/documentation/19.0/applications/inventory_and_mrp/purchase.html}{Purchase app documentation}.

\textbf{Manufacturing order.} A manufacturing order, or MO, records an internal production commitment. In ERP-Bench, manufacturing orders consume components, use workcenter capacity, and produce finished goods or subassemblies. Odoo documents manufacturing orders in the \href{https://www.odoo.com/documentation/19.0/applications/inventory_and_mrp/manufacturing.html}{Manufacturing app documentation}.

\textbf{Bill of materials.} A bill of materials, or BOM, is the recipe for making a product. It specifies which components and quantities are required to manufacture one unit. Odoo documents BOMs in the \href{https://www.odoo.com/documentation/19.0/applications/inventory_and_mrp/manufacturing/basic_setup/bill_configuration.html}{bill of materials documentation}.

\textbf{Workcenter.} A workcenter is a production resource with limited capacity. ERP-Bench uses workcenters to represent machines, stations, or production lines that constrain how much can be manufactured before a deadline. Odoo documents workcenters in the \href{https://www.odoo.com/documentation/19.0/applications/inventory_and_mrp/manufacturing/advanced_configuration/using_work_centers.html}{work centers documentation}.

\textbf{Vendor pricelist.} A vendor pricelist records a supplier offer for a product. The offer can include price, lead time, minimum order quantity, and other purchase constraints. Odoo documents vendor pricelists in the \href{https://www.odoo.com/documentation/19.0/applications/inventory_and_mrp/purchase/products/pricelist.html}{vendor pricelist documentation}.

\textbf{Route.} A route describes how Odoo supplies or moves a product. ERP-Bench tasks use buy routes, manufacture routes, or both. Odoo documents routes in the \href{https://www.odoo.com/documentation/19.0/applications/inventory_and_mrp/inventory/shipping_receiving/daily_operations/use_routes.html}{routes and push or pull rules documentation}.

\textbf{Invoice and downpayment.} An invoice records what the customer owes. A downpayment invoice records a required partial payment before the order is fully delivered. Odoo documents invoices in the \href{https://www.odoo.com/documentation/19.0/applications/finance/accounting/customer_invoices.html}{customer invoices documentation} and downpayments in the \href{https://www.odoo.com/documentation/19.0/applications/sales/sales/invoicing/down_payment.html}{down payments documentation}.

\textbf{Origin field.} The origin field is the traceability field on purchase and manufacturing records. ERP-Bench checks that new POs and MOs reference the sales orders they serve.

\textbf{Internal notes.} Internal notes are comments stored on Odoo records such as vendors, customers, and products. ERP-Bench uses these notes to encode operational policies that agents must read from the ERP state.

\section{Dataset, Scope, and Limitations}\label{app:dataset}

\subsection{Benchmark Scope}\label{b.1-benchmark-scope}

ERP-Bench evaluates long-horizon procurement and manufacturing work in a production-grade ERP. The tasks require agents to read persistent business state, create valid sales, purchase, manufacturing, and invoice records, satisfy operational constraints, and optimize measurable business objectives.

\subsection{Scope Boundaries}\label{b.2-scope-boundaries}

ERP-Bench focuses on explicit procurement and manufacturing decisions in Odoo. Out-of-scope workflows include free-form negotiation, human escalation, hidden organization politics, and tasks whose success requires evidence outside the final ERP state. The release treats Odoo behavior and module configuration as fixed during evaluation.

The benchmark boundary follows the formalization boundary. Anchor certifies optima for workflows encoded as terminal-state business logic in Odoo, so ERP-Bench measures work that can be expressed in that form.

\subsection{Dataset Summary}\label{b.3-dataset-summary}

The release contains 300 tasks generated with seed \texttt{20260506} under dataset version \texttt{2026-05-diverse-v5}.

\begin{longtable}[]{@{}lr@{}}
\toprule\noalign{}
Metric & Value \\
\midrule\noalign{}
\endhead
\bottomrule\noalign{}
\endlastfoot
Tasks & 300 \\
Workflow patterns & 29 \\
Easy tasks & 50 \\
Medium tasks & 166 \\
Hard tasks & 84 \\
Tasks with BOMs in the seeded ERP state & 290 \\
Tasks with workcenters in the seeded ERP state & 218 \\
Tasks whose optimal plan uses purchase orders & 268 \\
Tasks whose optimal plan uses manufacturing orders & 214 \\
Tasks with pre-existing sales orders & 105 \\
Tasks with pre-existing purchase orders & 28 \\
Tasks with pre-existing manufacturing orders & 20 \\
\end{longtable}

\subsection{Objective Mix}\label{b.4-objective-mix}

\begin{longtable}[]{@{}lr@{}}
\toprule\noalign{}
Objective kind & Tasks \\
\midrule\noalign{}
\endhead
\bottomrule\noalign{}
\endlastfoot
Minimum new spend & 183 \\
Vendor consolidation & 34 \\
Capacity preservation & 31 \\
Repair plan & 28 \\
Constraint only & 24 \\
\end{longtable}

Most tasks minimize new spend after satisfying the operational constraints. Vendor-consolidation tasks keep the same procurement feasibility problem and prefer fewer suppliers. Capacity-preservation tasks prefer plans that leave more factory capacity available after the customer orders are covered. Repair-plan tasks start from a broken plan and prefer small valid changes to the existing state. Constraint-only tasks grade feasibility without adding an economic tie-breaker.

\subsection{Pattern Inventory and Business Map}\label{b.5-pattern-inventory-and-business-map}

Each task pattern has a stable generator ID and a business-facing name; the tables below pair them with the planning structure a human sees.

\subsubsection{Buy and Intake Patterns}\label{b.5.1-buy-and-intake-patterns}

\begin{longtable}[]{@{}
  >{\raggedright\arraybackslash}p{0.03\linewidth}
  >{\raggedright\arraybackslash}p{0.26\linewidth}
  >{\raggedright\arraybackslash}p{0.18\linewidth}
  >{\raggedleft\arraybackslash}p{0.05\linewidth}
  >{\raggedright\arraybackslash}p{0.36\linewidth}@{}}
\toprule\noalign{}
No. & Generator pattern & Business name & Tasks & Planning structure \\
\midrule\noalign{}
\endhead
\bottomrule\noalign{}
\endlastfoot
1 & \texttt{01\_buy\_only\_baseline} & Routine replenishment cycle & 8 & Customer demand can be covered with warehouse stock plus supplier purchases. No factory planning, screening rule, or special billing step is present. \\
2 & \texttt{02\_buy\_only\_immediate\_invoicing} & Buy and bill immediately & 8 & The agent buys finished goods and invoices customers in the same cycle. This adds an order-to-cash step after sourcing. \\
3 & \texttt{03\_buy\_only\_fixed\_downpayment} & Buy with a fixed deposit & 8 & Large purchases require a fixed upfront payment before the rest of the billing flow. The cash-collection rule stays fixed across order sizes. \\
4 & \texttt{04\_buy\_only\_percentage\_downpayment} & Buy with a percentage deposit & 8 & Large purchases require a deposit that scales with order value. Deposit math changes with the order mix. \\
5 & \texttt{05\_screened\_buy\_only\_all\_seeded} & Screen a preloaded backlog before buying & 10 & All requests begin as draft ERP orders. The planner cancels requests that violate intake policy before sourcing accepted demand. \\
6 & \texttt{06\_screened\_buy\_only\_mixed\_seeded} & Screen mixed intake before buying & 11 & Some requests are already in ERP and others appear only in the task brief. The planner applies one intake policy across both channels. \\
7 & \texttt{07\_screened\_buy\_only\_mixed\_seeded\_invoicing} & Screen, buy, and invoice & 11 & Orders are filtered by policy, fulfilled through suppliers, and invoiced. Intake governance, sourcing, and billing happen in one flow. \\
8 & \texttt{26\_buy\_only\_net\_30\_no\_adjacent\_data} & Lean-context purchasing cycle & 10 & A normal buy-side fulfillment task has fewer surrounding clues and supporting records. The sourcing plan is straightforward and the context is sparse. \\
\end{longtable}

\subsubsection{Production and Fulfillment Patterns}\label{b.5.2-production-and-fulfillment-patterns}

\begin{longtable}[]{@{}
  >{\raggedright\arraybackslash}p{0.03\linewidth}
  >{\raggedright\arraybackslash}p{0.26\linewidth}
  >{\raggedright\arraybackslash}p{0.18\linewidth}
  >{\raggedleft\arraybackslash}p{0.05\linewidth}
  >{\raggedright\arraybackslash}p{0.36\linewidth}@{}}
\toprule\noalign{}
No. & Generator pattern & Business name & Tasks & Planning structure \\
\midrule\noalign{}
\endhead
\bottomrule\noalign{}
\endlastfoot
9 & \texttt{08\_single\_bom\_single\_workcenter} & One-line make-or-buy plan & 11 & Inventory is short, and the business can buy finished goods or build internally on one main line. One bottleneck line largely determines the answer. \\
10 & \texttt{09\_single\_bom\_lowest\_cost} & Lowest-cost sourcing route & 11 & Multiple supply routes are feasible. The objective selects the cheapest workable mix. \\
11 & \texttt{10\_single\_bom\_split\_by\_capacity} & Capacity-smoothing production plan & 11 & The lowest-spend route can overload the main line, so work must be split across available capacity. The plan protects bottlenecks as part of fulfillment. \\
12 & \texttt{11\_restricted\_subassembly\_qualified\_workcenters} & Restricted-module manufacturing plan & 10 & A critical module has to be built internally, and only qualified stations can make it. The binding constraint sits on the restricted subassembly. \\
13 & \texttt{12\_single\_subassembly\_lowest\_cost} & Two-stage build with cheapest branch & 11 & The product has one intermediate build step before final assembly. Cost is optimized across a deeper build tree. \\
14 & \texttt{13\_single\_subassembly\_qualified\_workcenters} & Two-stage build with qualification rules & 11 & One intermediate build step has station-qualification rules. Capability and certification shape the feasible production plan. \\
15 & \texttt{14\_single\_subassembly\_shared\_overflow\_capacity} & Shared-overflow manufacturing plan & 11 & Several production steps can use the same backup capacity pool when primary capacity fills. Backup capacity consumed in one step reduces options elsewhere. \\
16 & \texttt{15\_parallel\_subassemblies\_branch\_assigned} & Parallel branch production & 10 & Different subassemblies can be built at the same time on designated lines. The planner coordinates parallel workstreams. \\
17 & \texttt{16\_serial\_subassemblies\_branch\_assigned} & Sequential build-chain production & 10 & The product requires multiple steps in a strict order before final assembly can finish. Dependency timing drives the plan. \\
18 & \texttt{17\_shared\_component\_subassemblies\_branch\_assigned} & Shared-component production & 11 & Several branches need the same common part or subassembly. Different workstreams compete for the same scarce input. \\
19 & \texttt{18\_manufacture\_only\_policy\_forbidden} & Factory-only by policy & 11 & Company policy forbids buying finished goods for this cycle. Internal production is the available fulfillment path. \\
20 & \texttt{19\_manufacture\_only\_no\_buy\_route} & Factory-only by system design & 11 & The ERP setup has no buy route for the finished item. The constraint comes from product configuration. \\
21 & \texttt{20\_manufacture\_only\_no\_available\_vendors} & Factory-only by market reality & 12 & Finished-goods vendors are unavailable. Market availability forces internal production. \\
22 & \texttt{21\_single\_bom\_lowest\_cost\_screened\_mixed\_seeded} & Screened backlog with vendor consolidation & 12 & Orders are screened first, then accepted demand is covered using as few suppliers as possible. Intake governance and supplier rationalization interact. \\
23 & \texttt{22\_single\_bom\_split\_by\_capacity\_invoicing} & Capacity-smoothing plan with billing & 12 & Demand is split across lines to protect capacity, then invoiced correctly. Production planning feeds downstream billing execution. \\
24 & \texttt{23\_restricted\_subassembly\_qualified\_workcenters\_screened\_all\_seeded} & Screened restricted-module plan & 11 & Preloaded draft orders are screened, and accepted demand depends on a restricted internal module. Backlog triage meets capability-constrained manufacturing. \\
25 & \texttt{24\_single\_subassembly\_shared\_overflow\_capacity\_screened\_invoicing} & Screened shared-overflow plan with invoicing & 11 & Rule-compliant orders proceed, then compete for shared backup capacity and require invoicing. Intake, capacity sharing, and billing interact. \\
26 & \texttt{25\_shared\_component\_subassemblies\_branch\_assigned\_screened\_all\_seeded} & Screened shared-component plan & 11 & Preloaded orders are screened, and the accepted set competes for shared intermediate parts. The plan balances shared materials after intake triage. \\
\end{longtable}

\subsubsection{Disruption Recovery Patterns}\label{b.5.3-disruption-recovery-patterns}

\begin{longtable}[]{@{}
  >{\raggedright\arraybackslash}p{0.03\linewidth}
  >{\raggedright\arraybackslash}p{0.26\linewidth}
  >{\raggedright\arraybackslash}p{0.18\linewidth}
  >{\raggedleft\arraybackslash}p{0.05\linewidth}
  >{\raggedright\arraybackslash}p{0.36\linewidth}@{}}
\toprule\noalign{}
No. & Generator pattern & Business name & Tasks & Planning structure \\
\midrule\noalign{}
\endhead
\bottomrule\noalign{}
\endlastfoot
27 & \texttt{repair\_plan\_easy} & Supplier cancellation rescue & 8 & A confirmed supplier backs out after customer commitments already exist, and no in-house production fallback is available. The task recovers from a broken purchase plan. \\
28 & \texttt{repair\_plan\_medium} & Supplier cancellation rescue with production fallback & 10 & A supplier fails, and some of the gap can be recovered through internal production. The planner rebalances existing purchase and manufacturing commitments. \\
29 & \texttt{repair\_plan\_hard} & Line-outage recovery plan & 10 & The original production plan breaks when the main assembly line goes down mid-cycle. The planner reroutes work to alternate lines and rebuilds the schedule under reduced capacity. \\
\end{longtable}

Stable task names include the scenario number, difficulty, and generator pattern. This makes per-pattern result tables reproducible from the directory names.

\subsection{ERP State Scale}\label{b.6-erp-state-scale}

\begin{longtable}[]{@{}
  >{\raggedright\arraybackslash}p{0.1467\linewidth}
  >{\raggedleft\arraybackslash}p{0.1467\linewidth}
  >{\raggedleft\arraybackslash}p{0.1467\linewidth}
  >{\raggedleft\arraybackslash}p{0.1467\linewidth}
  >{\raggedleft\arraybackslash}p{0.1467\linewidth}
  >{\raggedleft\arraybackslash}p{0.1467\linewidth}@{}}
\toprule\noalign{}
Difficulty & Customers mean & Vendors mean & Products mean & BOMs mean & Workcenters mean \\
\midrule\noalign{}
\endhead
\bottomrule\noalign{}
\endlastfoot
Easy & 44.00 & 43.60 & 41.00 & 8.00 & 0.00 \\
Medium & 58.78 & 70.68 & 55.60 & 11.01 & 1.75 \\
Hard & 67.75 & 69.29 & 56.21 & 12.38 & 3.29 \\
\end{longtable}

The seeded ERP state includes many more customers, vendors, and products than the task directly mentions. Agents must find the relevant records, use task-specific stock and supplier facts, and leave unrelated business records unchanged.

\subsection{Product Domains and Routes}\label{b.7-product-domains-and-routes}

\begin{longtable}[]{@{}lr@{}}
\toprule\noalign{}
Product category of target product & Tasks \\
\midrule\noalign{}
\endhead
\bottomrule\noalign{}
\endlastfoot
Power Equipment & 53 \\
Industrial Equipment & 41 \\
Office Systems & 34 \\
Material Handling Equipment & 34 \\
Building Systems & 33 \\
Safety Systems & 22 \\
Commercial Fixtures & 21 \\
Data Center Equipment & 19 \\
Industrial Automation & 11 \\
Utility Equipment & 11 \\
Cleanroom Equipment & 11 \\
Laboratory Equipment & 10 \\
\end{longtable}

\begin{longtable}[]{@{}lr@{}}
\toprule\noalign{}
Target product route & Tasks \\
\midrule\noalign{}
\endhead
\bottomrule\noalign{}
\endlastfoot
Buy & 82 \\
Buy and manufacture & 207 \\
Manufacture & 11 \\
\end{longtable}

Product categories change the product names, adjacent catalog state, and surface vocabulary. Routes determine the available action space. Buy-route tasks require purchase orders for external supply. Manufacture-route tasks require manufacturing orders, components, and workcenter capacity. Mixed-route tasks let the agent choose between external supply and internal production.

\section{Task Generation}\label{app:ranges}

\subsection{Generator Inputs}\label{c.1-generator-inputs}

The 300-task recipe is \texttt{diverse\_300\_dataset.toml}. It enumerates 29 scenario TOMLs with per-pattern counts. Each scenario TOML applies schema defaults from \texttt{config.py}. The recipe and scenario TOMLs ship under \texttt{procurement/examples/} in the release.

The tables below define every recipe field and then show six representative scenarios that exercise the full field surface. Pattern numbers match Appendix B.5.

\textbf{Field notes.}

\begin{longtable}[]{@{}
  >{\raggedright\arraybackslash}p{0.4400\linewidth}
  >{\raggedright\arraybackslash}p{0.4400\linewidth}@{}}
\toprule\noalign{}
Field & Meaning \\
\midrule\noalign{}
\endhead
\bottomrule\noalign{}
\endlastfoot
\texttt{Customers} & Range for the number of task-relevant customer orders sampled into the scenario \\
\texttt{Deadline\ days} & Range for customer due dates, measured in days from task start \\
\texttt{Demand} & Range for units requested by each task-relevant customer \\
\texttt{Stock\ ratio} & Starting finished-goods stock as a fraction of total sampled demand \\
\texttt{Finished\ vendors} & Target range for finished-goods suppliers, followed by the hard vendor limit \\
\texttt{Vendor\ cap\ ratio} & Supplier capacity range, expressed relative to total demand \\
\texttt{Tightness} & Sampler pressure knob. Larger values sample tighter supply and capacity conditions \\
\texttt{Components} & Range for component products generated for manufacturing scenarios \\
\texttt{Component\ vendor\ cats} & Range for supplier categories used when generating component vendors \\
\texttt{Capacity\ ratio} & Assembly capacity range, expressed relative to total demand before the integer buffer \\
\texttt{Intake\ mode} & Whether every request is accepted or draft orders are screened by policy \\
\texttt{Seeded\ coverage} & Whether screened requests are all preloaded in ERP or split between ERP drafts and the task brief \\
\texttt{Reject\ rules} & Intake rules that mark a customer request for cancellation \\
\texttt{Invoice\ policy} & Required payment term and downpayment policy for customer invoices \\
\texttt{Repair} & Disruption seeded into the initial state for repair-plan tasks \\
\texttt{Adjacent\ data} & Whether unrelated customers, vendors, products, and records are added to the ERP state \\
\end{longtable}

\textbf{Representative scenarios.} The six scenarios below span the difficulty levels, route types, objective kinds, intake modes, invoice policies, BOM structures, and repair disruptions used across the 300-task recipe. The remaining 23 scenarios reuse these same field values in different combinations.

\textbf{Core recipe fields.}

\begin{longtable}[]{@{}
  >{\raggedright\arraybackslash}p{0.1258\linewidth}
  >{\raggedright\arraybackslash}p{0.1258\linewidth}
  >{\raggedleft\arraybackslash}p{0.1258\linewidth}
  >{\raggedright\arraybackslash}p{0.1258\linewidth}
  >{\raggedright\arraybackslash}p{0.1258\linewidth}
  >{\raggedright\arraybackslash}p{0.1258\linewidth}
  >{\raggedright\arraybackslash}p{0.1258\linewidth}@{}}
\toprule\noalign{}
No. & Config & Tasks & Difficulty & Route & Objective & Domain \\
\midrule\noalign{}
\endhead
\bottomrule\noalign{}
\endlastfoot
1 & \texttt{01\_buy\_only\_baseline} & 8 & \texttt{easy} & \texttt{buy\_only} & \texttt{constraint\_only} & \texttt{sound\_panels} \\
4 & \texttt{04\_buy\_only\_percentage\_downpayment} & 8 & \texttt{easy} & \texttt{buy\_only} & \texttt{min\_new\_spend} & \texttt{workstations} \\
7 & \texttt{07\_screened\_buy\_only\_mixed\_seeded\_invoicing} & 11 & \texttt{medium} & \texttt{buy\_only} & \texttt{min\_new\_spend} & \texttt{hvac\_units} \\
15 & \texttt{15\_parallel\_subassemblies\_branch\_assigned} & 10 & \texttt{hard} & \texttt{buy\_and\_manufacture} & \texttt{capacity\_preservation} & \texttt{battery\_packs} \\
21 & \texttt{21\_single\_bom\_lowest\_cost\_screened\_mixed\_seeded} & 12 & \texttt{medium} & \texttt{buy\_and\_manufacture} & \texttt{vendor\_consolidation} & \texttt{packaging\_machines} \\
29 & \texttt{repair\_plan\_hard} & 10 & \texttt{hard} & \texttt{buy\_and\_manufacture} & \texttt{repair\_plan} & \texttt{solar\_panels} \\
\end{longtable}

\textbf{Demand and supply ranges.}

\begin{longtable}[]{@{}
  >{\raggedright\arraybackslash}p{0.1100\linewidth}
  >{\raggedright\arraybackslash}p{0.1100\linewidth}
  >{\raggedright\arraybackslash}p{0.1100\linewidth}
  >{\raggedright\arraybackslash}p{0.1100\linewidth}
  >{\raggedright\arraybackslash}p{0.1100\linewidth}
  >{\raggedright\arraybackslash}p{0.1100\linewidth}
  >{\raggedright\arraybackslash}p{0.1100\linewidth}
  >{\raggedleft\arraybackslash}p{0.1100\linewidth}@{}}
\toprule\noalign{}
No. & Customers & Deadline days & Demand & Stock ratio & Finished vendors & Vendor cap ratio & Tightness \\
\midrule\noalign{}
\endhead
\bottomrule\noalign{}
\endlastfoot
1 & \texttt{{[}4,\ 4{]}} & \texttt{{[}5,\ 9{]}} & \texttt{{[}1,\ 11{]}} & \texttt{{[}0.75,\ 0.92{]}} & \texttt{{[}3,\ 3{]}}, limit 4 & \texttt{{[}0.40,\ 0.90{]}} & 0.25 \\
4 & \texttt{{[}4,\ 4{]}} & \texttt{{[}5,\ 9{]}} & \texttt{{[}1,\ 11{]}} & \texttt{{[}0.75,\ 0.92{]}} & \texttt{{[}3,\ 3{]}}, limit 4 & \texttt{{[}0.40,\ 0.90{]}} & 0.25 \\
7 & \texttt{{[}8,\ 10{]}} & \texttt{{[}8,\ 14{]}} & \texttt{{[}14,\ 25{]}} & \texttt{{[}0.38,\ 0.52{]}} & \texttt{{[}4,\ 5{]}}, limit 6 & \texttt{{[}0.10,\ 0.36{]}} & 0.55 \\
15 & \texttt{{[}20,\ 32{]}} & \texttt{{[}8,\ 13{]}} & \texttt{{[}18,\ 31{]}} & \texttt{{[}0.28,\ 0.42{]}} & \texttt{{[}5,\ 6{]}}, limit 8 & \texttt{{[}0.07,\ 0.26{]}} & 0.72 \\
21 & \texttt{{[}8,\ 10{]}} & \texttt{{[}8,\ 14{]}} & \texttt{{[}14,\ 25{]}} & \texttt{{[}0.38,\ 0.52{]}} & \texttt{{[}4,\ 5{]}}, limit 6 & \texttt{{[}0.10,\ 0.36{]}} & 0.55 \\
29 & \texttt{{[}10,\ 12{]}} & \texttt{{[}8,\ 12{]}} & \texttt{{[}15,\ 26{]}} & \texttt{{[}0.04,\ 0.10{]}} & \texttt{{[}4,\ 5{]}}, limit 6 & \texttt{{[}0.07,\ 0.24{]}} & 0.62 \\
\end{longtable}

\textbf{Intake and billing fields.}

\begin{longtable}[]{@{}
  >{\raggedright\arraybackslash}p{0.1760\linewidth}
  >{\raggedright\arraybackslash}p{0.1760\linewidth}
  >{\raggedright\arraybackslash}p{0.1760\linewidth}
  >{\raggedright\arraybackslash}p{0.1760\linewidth}
  >{\raggedright\arraybackslash}p{0.1760\linewidth}@{}}
\toprule\noalign{}
No. & Intake mode & Seeded coverage & Reject rules & Invoice policy \\
\midrule\noalign{}
\endhead
\bottomrule\noalign{}
\endlastfoot
1 & \texttt{accept\_all\_orders} & omitted & omitted & omitted \\
4 & \texttt{accept\_all\_orders} & omitted & omitted & \texttt{net\_30}, \texttt{percentage} 40 over 2000 \\
7 & \texttt{screen\_orders\_by\_policy} & \texttt{mixed\_seeded\_and\_prompt\_only} & \texttt{quantity\_outside\_window}, \texttt{lead\_time\_below\_minimum} & \texttt{net\_30} \\
15 & \texttt{accept\_all\_orders} & omitted & omitted & omitted \\
21 & \texttt{screen\_orders\_by\_policy} & \texttt{mixed\_seeded\_and\_prompt\_only} & \texttt{budget\_below\_list\_price}, \texttt{lead\_time\_below\_minimum} & omitted \\
29 & \texttt{accept\_all\_orders} & omitted & omitted & omitted \\
\end{longtable}

Other recipe rows reuse the same field options. The \texttt{immediate} invoice policy appears in \texttt{02\_buy\_only\_immediate\_invoicing} and \texttt{22\_single\_bom\_split\_by\_capacity\_invoicing}. A \texttt{fixed\_amount} downpayment appears in \texttt{03\_buy\_only\_fixed\_downpayment}. The \texttt{all\_seeded\_orders} seeded coverage and the \texttt{budget\_below\_list\_price} reject rule appear in \texttt{05\_screened\_buy\_only\_all\_seeded}, \texttt{23\_restricted\_subassembly\_qualified\_workcenters\_screened\_all\_seeded}, and \texttt{25\_shared\_component\_subassemblies\_branch\_assigned\_screened\_all\_seeded}.

\textbf{Manufacturing and repair fields.}

{\footnotesize
\setlength{\tabcolsep}{2pt}
\begin{longtable}[]{@{}
  >{\raggedright\arraybackslash}p{0.0978\linewidth}
  >{\raggedright\arraybackslash}p{0.0978\linewidth}
  >{\raggedright\arraybackslash}p{0.0978\linewidth}
  >{\raggedright\arraybackslash}p{0.0978\linewidth}
  >{\raggedright\arraybackslash}p{0.0978\linewidth}
  >{\raggedright\arraybackslash}p{0.0978\linewidth}
  >{\raggedright\arraybackslash}p{0.0978\linewidth}
  >{\raggedright\arraybackslash}p{0.0978\linewidth}
  >{\raggedright\arraybackslash}p{0.0978\linewidth}@{}}
\toprule\noalign{}
No. & Components & Component vendor cats & BOM structure & Workcenters & Workcenter path & Capacity ratio & Repair & Adjacent data \\
\midrule\noalign{}
\endhead
\bottomrule\noalign{}
\endlastfoot
1 & omitted & \texttt{{[}1,\ 1{]}} & omitted & omitted & omitted & omitted & omitted & \texttt{true} \\
4 & omitted & \texttt{{[}1,\ 1{]}} & omitted & omitted & omitted & omitted & omitted & \texttt{true} \\
7 & omitted & \texttt{{[}1,\ 1{]}} & omitted & omitted & omitted & omitted & omitted & \texttt{true} \\
15 & \texttt{{[}8,\ 10{]}} & \texttt{{[}3,\ 3{]}} & \texttt{parallel\_subassemblies} & \texttt{3} & \texttt{branch\_assigned\_workcenters} & \texttt{{[}0.20,\ 0.34{]}} & omitted & \texttt{true} \\
21 & \texttt{{[}5,\ 7{]}} & \texttt{{[}3,\ 3{]}} & \texttt{single\_bom} & \texttt{3} & \texttt{lowest\_cost} & \texttt{{[}0.24,\ 0.46{]}} & omitted & \texttt{true} \\
29 & \texttt{{[}4,\ 6{]}} & \texttt{{[}3,\ 3{]}} & \texttt{single\_bom} & \texttt{3} & \texttt{split\_by\_capacity} & \texttt{{[}0.18,\ 0.32{]}} & \texttt{workcenter\_outage} & \texttt{true} \\
\end{longtable}
}

The full recipe also uses the \texttt{single\_subassembly}, \texttt{serial\_subassemblies}, \texttt{restricted\_subassembly}, and \texttt{shared\_component\_subassemblies} BOM structures, the \texttt{qualified\_workcenters} and \texttt{shared\_overflow\_capacity} workcenter paths, and the \texttt{supplier\_cancellation} repair disruption. Adjacent data is enabled in 28 of 29 scenarios; \texttt{26\_buy\_only\_net\_30\_no\_adjacent\_data} is the lean-context exception that disables it.

\subsection{Rejection and Resampling}\label{c.2-rejection-and-resampling}

Anchor generated 300 accepted tasks after rejecting 732 sampled parameter sets. Rejection happens before release, so unsatisfiable parameterizations never become benchmark tasks.

\begin{longtable}[]{@{}lr@{}}
\toprule\noalign{}
Generation metric & Value \\
\midrule\noalign{}
\endhead
\bottomrule\noalign{}
\endlastfoot
Accepted tasks & 300 \\
Wall time & 656.76s \\
Total CP-SAT solve time & 651.93s \\
Average solve time per accepted task & 2.173s \\
Discarded parameter sets & 732 \\
Pre-solver discards & 275 \\
Solver discards & 457 \\
Tasks with at least one resample & 155 \\
Tasks accepted without resampling & 145 \\
\end{longtable}

Anchor rejects infeasible source instances at generation time. Rejected samples never become task directories, so accepted task directories have a solver-certified feasible terminal state before any agent evaluation begins.

\subsection{Solver Performance by Difficulty}\label{c.3-solver-performance-by-difficulty}

{\footnotesize
\setlength{\tabcolsep}{2pt}
\begin{longtable}[]{@{}
  >{\raggedright\arraybackslash}p{0.0800\linewidth}
  >{\raggedleft\arraybackslash}p{0.0800\linewidth}
  >{\raggedleft\arraybackslash}p{0.0800\linewidth}
  >{\raggedleft\arraybackslash}p{0.0800\linewidth}
  >{\raggedleft\arraybackslash}p{0.0800\linewidth}
  >{\raggedleft\arraybackslash}p{0.0800\linewidth}
  >{\raggedleft\arraybackslash}p{0.0800\linewidth}
  >{\raggedleft\arraybackslash}p{0.0800\linewidth}
  >{\raggedleft\arraybackslash}p{0.0800\linewidth}
  >{\raggedleft\arraybackslash}p{0.0800\linewidth}
  >{\raggedleft\arraybackslash}p{0.0800\linewidth}@{}}
\toprule\noalign{}
Difficulty & Tasks & Avg solve & Median & P95 & Max & Avg vars & Avg tie-break vars & Avg calls & Discards & Resampled tasks \\
\midrule\noalign{}
\endhead
\bottomrule\noalign{}
\endlastfoot
Easy & 50 & 0.044s & 0.008s & 0.288s & 0.347s & 19.76 & 18.20 & 16.36 & 253 & 20 \\
Medium & 166 & 0.951s & 0.031s & 2.842s & 20.863s & 78.57 & 77.45 & 4.69 & 243 & 79 \\
Hard & 84 & 5.856s & 0.100s & 31.984s & 100.735s & 96.67 & 95.20 & 19.02 & 236 & 56 \\
\end{longtable}
}

Hard tasks combine richer sourcing, manufacturing, capacity, screening, and repair constraints, which increases solve time. The long tail of hard solve times comes from resampling difficult parameter settings until the solver certifies a valid task.

\subsection{Solver Status Accounting}\label{c.4-solver-status-accounting}

\begin{longtable}[]{@{}lrrrr@{}}
\toprule\noalign{}
Phase & Calls & Optimal & Infeasible & Unknown \\
\midrule\noalign{}
\endhead
\bottomrule\noalign{}
\endlastfoot
Feasibility & 24 & 24 & 0 & 0 \\
Primary objective & 1,324 & 1,120 & 204 & 0 \\
Primary retry & 204 & 0 & 204 & 0 \\
Spend secondary & 468 & 468 & 0 & 0 \\
Lexicographic fixed-search & 1,144 & 1,113 & 0 & 31 \\
Lexicographic fixed-search retry & 31 & 3 & 0 & 28 \\
Total & 3,195 & 2,728 & 408 & 59 \\
\end{longtable}

Unknown statuses occur in tie-break phases for some sampled candidates. The accepted task still has a certified feasible solution and objective value. The fixed-search retry accounts for deterministic artifact selection.

\section{Artifact Format}\label{app:artifact-format}

\subsection{Task Bundle Layout}\label{d.1-task-bundle-layout}

Each task directory contains a complete executable benchmark bundle.

\begin{longtable}[]{@{}
  >{\raggedright\arraybackslash}p{0.4400\linewidth}
  >{\raggedright\arraybackslash}p{0.4400\linewidth}@{}}
\toprule\noalign{}
Path & Role \\
\midrule\noalign{}
\endhead
\bottomrule\noalign{}
\endlastfoot
\texttt{task.toml} & Metadata, timeouts, category, difficulty, objective kind, pattern, and seed \\
\texttt{instruction.md} & Natural-language task brief, ERP access instructions, and tool guidance \\
\texttt{environment/scenario\_data.json} & Structured sampled state used to seed Odoo \\
\texttt{environment/setup\_scenario.py} & Script that writes the initial ERP state \\
\texttt{environment/entrypoint.sh} & Container entrypoint \\
\texttt{environment/odoo.conf} & Odoo configuration \\
\texttt{environment/Dockerfile} & Per-task environment build \\
\texttt{solution/optimal\_plan.json} & Solver-certified plan, allocations, objective values, and accounting summary \\
\texttt{solution/solver.py} & Executable oracle writer emitted from the same solved plan \\
\texttt{solution/solve.sh} & Oracle entrypoint that runs the writer against Odoo \\
\texttt{tests/checks.py} & Terminal-state verifier \\
\texttt{tests/test.sh} & Verifier execution entrypoint \\
\end{longtable}

A reviewer can read \texttt{instruction.md} for the task brief, inspect \texttt{scenario\_data.json} for the seeded state, run \texttt{solve.sh} to replay the solver-certified plan, and run \texttt{test.sh} to score any terminal state --- all without touching the original generator.

\begin{Verbatim}[breaklines=true,breakanywhere=true,fontsize=\footnotesize]
flowchart LR
    Setup["environment/setup_scenario.py"] --> Seeded["Seeded Odoo state"]
    Seeded --> Agent["Agent trial"]
    Seeded --> Oracle["Oracle replay<br/>(solution/solve.sh)"]
    Seeded --> Noop["No-op replay"]
    Agent --> Terminal["Terminal Odoo state"]
    Oracle --> Terminal
    Noop --> Terminal
    Terminal --> Verifier["tests/test.sh<br/>(checks.py)"]
    Verifier --> Logs["reward.json, spend.json,<br/>optimality.json, rule_results.tsv"]
\end{Verbatim}

\texttt{setup\_scenario.py} writes \texttt{scenario\_data.json} into Odoo to produce the seeded state. Agent trials, oracle replay, and no-op replay all start from that seeded state and converge on a terminal Odoo state. The same verifier grades every terminal state and emits the same score artifacts.

\subsection{Representative Task Bundle}\label{d.2-representative-task-bundle}

This section gives raw excerpts from \texttt{2000\_easy\_01\_buy\_only\_baseline}. The JSON excerpts keep the task-relevant records and omit adjacent customers, vendors, products, and BOMs. The example is a small buy-only task, so the full business problem fits in four customer orders, one product, four vendor offers, and one oracle purchase order.

\texttt{task.toml}:

\begin{Verbatim}[breaklines=true,breakanywhere=true,fontsize=\footnotesize]
version = "1.0"

[metadata]
scenario_number = 2000
name = "Sound Privacy Panels"
difficulty = "easy"
category = "erp"
objective_kind = "constraint_only"
task_pattern = "01_buy_only_baseline"
tags = ["odoo", "erp-bench", "procurement"]
seed = 8014130904848695607

[verifier]
timeout_sec = 300.0

[agent]
timeout_sec = 3600.0

[environment]
build_timeout_sec = 600.0
cpus = 3
memory_mb = 4096
storage_mb = 2048
\end{Verbatim}

\texttt{instruction.md} excerpt:

\begin{Verbatim}[breaklines=true,breakanywhere=true,fontsize=\footnotesize]
Supply plans must be created to fulfill each outstanding Sound Privacy Panels customer order listed.

- Coral Clinics: 4 units due in 6 days (pretax budget cap $1,922)
- Ironside Exchange: 7 units due in 7 days (pretax budget cap $3,318)
- Element Productions: 12 units due in 8 days (pretax budget cap $5,425)
- Spire Studios East: 9 units due in 8 days (pretax budget cap $4,363)

On-hand finished stock covers 27 units. There is no in-house manufacturing route for the finished product, so any shortfall has to come from vendors.

Across all units covered through new purchasing or manufacturing, keep portfolio-level new-spend margin at or above 28.7% at selling price.
\end{Verbatim}

\texttt{environment/scenario\_data.json} excerpt:

\begin{Verbatim}[breaklines=true,breakanywhere=true,fontsize=\footnotesize]
{
  "scenario_number": 2000,
  "name": "Sound Privacy Panels",
  "seed": 8014130904848695607,
  "customers": [
    {
      "name": "Coral Clinics",
      "ref": "r674891CF17_c01",
      "budget_dollars": 1921.97
    },
    {
      "name": "Ironside Exchange",
      "ref": "r674891CF17_c02",
      "budget_dollars": 3317.67
    },
    {
      "name": "Element Productions",
      "ref": "r674891CF17_c03",
      "budget_dollars": 5424.59
    },
    {
      "name": "Spire Studios East",
      "ref": "r674891CF17_c04",
      "budget_dollars": 4362.92
    }
  ],
  "products": [
    {
      "name": "Sound Privacy Panels",
      "code": "P674891CF17-SPP-001",
      "category": "Office Systems",
      "type": "consu",
      "list_price": 442.85,
      "standard_price": 304.86,
      "routes": ["buy"],
      "vendor_info": [
        {
          "partner_ref": "r674891CF17_q01",
          "delay": 6,
          "min_qty": 11.0,
          "max_qty": 31.0,
          "price": 278.72,
          "name": "Priority Shuttle lane (6d transit, 11-31 units)"
        },
        {
          "partner_ref": "r674891CF17_q02",
          "delay": 3,
          "min_qty": 3.0,
          "max_qty": 31.0,
          "price": 380.63,
          "name": "Express Air lane (3d transit, 3-31 units)"
        },
        {
          "partner_ref": "r674891CF17_q03",
          "delay": 18,
          "min_qty": 18.0,
          "max_qty": 31.0,
          "price": 143.53,
          "name": "Bulk Container lane (18d transit, 18-31 units)"
        },
        {
          "partner_ref": "r674891CF17_q04",
          "delay": 12,
          "min_qty": 9.0,
          "max_qty": 16.0,
          "price": 253.3,
          "name": "Scheduled Freight lane (12d transit, 9-16 units)"
        }
      ]
    }
  ],
  "stock_levels": [
    {
      "product_code": "P674891CF17-SPP-001",
      "warehouse_code": null,
      "quantity": 27.0,
      "location_type": "stock"
    }
  ],
  "existing_purchase_orders": [],
  "existing_manufacturing_orders": [],
  "existing_sales_orders": [],
  "repair_context": null,
  "system_parameters": [
    {
      "key": "saas_bench.min_margin_percent",
      "value": "28.7"
    }
  ]
}
\end{Verbatim}

\texttt{solution/optimal\_plan.json} excerpt:

\begin{Verbatim}[breaklines=true,breakanywhere=true,fontsize=\footnotesize]
{
  "objective_kind": "constraint_only",
  "objective_value": 0.0,
  "allocations": [
    {
      "ref": "r674891CF17_c01",
      "accepted": true,
      "action": "create_confirm_prompt",
      "stock_allocated": 0,
      "demand": 4,
      "fulfilled_demand": 4,
      "deadline": 6,
      "price": 442.85,
      "purchase_units": 4,
      "assembly_units": 0,
      "margin": 0.3706221068081743
    },
    {
      "ref": "r674891CF17_c02",
      "accepted": true,
      "action": "create_confirm_prompt",
      "stock_allocated": 0,
      "demand": 7,
      "fulfilled_demand": 7,
      "deadline": 7,
      "price": 442.85,
      "purchase_units": 7,
      "assembly_units": 0,
      "margin": 0.3706221068081743
    },
    {
      "ref": "r674891CF17_c03",
      "accepted": true,
      "action": "create_confirm_prompt",
      "stock_allocated": 12,
      "demand": 12,
      "fulfilled_demand": 12,
      "deadline": 8,
      "price": 442.85,
      "purchase_units": 0,
      "assembly_units": 0,
      "margin": 0.3115953483120695
    },
    {
      "ref": "r674891CF17_c04",
      "accepted": true,
      "action": "create_confirm_prompt",
      "stock_allocated": 9,
      "demand": 9,
      "fulfilled_demand": 9,
      "deadline": 8,
      "price": 442.85,
      "purchase_units": 0,
      "assembly_units": 0,
      "margin": 0.31159534831206964
    }
  ],
  "purchase_orders": [
    {
      "vendor_ref": "r674891CF17_q01",
      "quantity": 11,
      "unit_cost": 278.72,
      "total_cost": 3065.92,
      "lead_time": 6,
      "planned_arrival_days": 6,
      "product_code": "P674891CF17-SPP-001",
      "supply_role": "finished",
      "origin_customer_refs": [
        "r674891CF17_c01",
        "r674891CF17_c02"
      ]
    }
  ],
  "manufacturing_orders": [],
  "stock_report": [
    {
      "code": "P674891CF17-SPP-001",
      "label": "Sound Privacy Panels",
      "available": 27,
      "used": 21,
      "remaining": 6
    }
  ],
  "optimal_new_spend": 3065.92,
  "distinct_vendors_used": 1,
  "finished_purchase_spend": 3065.92,
  "expected_new_spend": 3065.92,
  "new_spend_units": 11,
  "revenue_from_new_spend": 4871.35,
  "new_spend_margin": 0.37062210680817437,
  "total_revenue": 14171.2,
  "total_cost": 9467.98,
  "margin": 0.33188579654510564
}
\end{Verbatim}

A reader can translate the raw bundle into the ERP action sequence below.

\begin{longtable}[]{@{}
  >{\raggedright\arraybackslash}p{0.4400\linewidth}
  >{\raggedright\arraybackslash}p{0.4400\linewidth}@{}}
\toprule\noalign{}
Bundle evidence & ERP action \\
\midrule\noalign{}
\endhead
\bottomrule\noalign{}
\endlastfoot
Four customer entries in \texttt{instruction.md} and \texttt{customers} & Create and confirm four sales orders for Sound Privacy Panels \\
\texttt{list\_price\ =\ 442.85} and customer budgets & Price each sales order at list price and keep each pretax order total under the customer budget \\
Deadlines of 6, 7, 8, and 8 days & Set commitment dates on the four sales orders \\
\texttt{stock\_report}: 27 available, 21 used, 6 remaining & Let stock cover the 12-unit Element order and the 9-unit Spire order \\
\texttt{purchase\_orders{[}0{]}}: vendor \texttt{r674891CF17\_q01}, quantity 11, delivery day 6 & Create and confirm one purchase order for the 11 units needed by Coral and Ironside \\
\texttt{origin\_customer\_refs}: Coral and Ironside & Put the generated sales order names in the purchase order origin field \\
\texttt{manufacturing\_orders}: empty & Create no manufacturing orders \\
\texttt{new\_spend\_margin\ =\ 0.3706} & The 11 purchased units clear the 28.7 percent new-spend margin rule \\
\end{longtable}

Raw \texttt{solution/solve.sh} excerpt:

\begin{Verbatim}[breaklines=true,breakanywhere=true,fontsize=\footnotesize]
set -eo pipefail

echo "=== Executing Optimal Solution ==="
echo "Scenario: 2000 - Sound Privacy Panels"

python3 /solution/solver.py

echo "Total revenue: $14171.20"
echo "Total cost: $9467.98"
echo "Required new-spend margin: 28.7%"
echo "New spend: $3065.92"
echo "Actual new-spend margin: 37.1%"
\end{Verbatim}

Abridged \texttt{solution/solver.py} excerpts; helper definitions and repeated sales metadata are elided:

\begin{Verbatim}[breaklines=true,breakanywhere=true,fontsize=\footnotesize]
SOLVER_PLAN = {
    "sales": [
        # Full entries also include task_order_ref values used as client_order_ref.
        {"ref": "r674891CF17_c01", "demand": 4, "deadline": 6, "price": 442.85},
        {"ref": "r674891CF17_c02", "demand": 7, "deadline": 7, "price": 442.85},
        {"ref": "r674891CF17_c03", "demand": 12, "deadline": 8, "price": 442.85},
        {"ref": "r674891CF17_c04", "demand": 9, "deadline": 8, "price": 442.85},
    ],
    "manufacturing": [],
    "purchases": [
        {
            "vendor_ref": "r674891CF17_q01",
            "product_code": "P674891CF17-SPP-001",
            "quantity": 11,
            "unit_cost": 278.72,
            "planned_arrival_days": 6,
            "supply_role": "finished",
            "origin_customer_refs": ["r674891CF17_c01", "r674891CF17_c02"],
        }
    ],
}
\end{Verbatim}

\begin{Verbatim}[breaklines=true,breakanywhere=true,fontsize=\footnotesize]
def execute(self):
    self._create_sale_orders()
    self._cancel_repair_artifacts()
    self._create_manufacturing_orders()
    self._create_purchase_orders()

def _create_confirmed_sale_order(self, sale_order_model, sale_order_line_model, step):
    sale_order_vals = {
        "partner_id": self._get_partner_id(step["ref"]),
        "commitment_date": self._scenario_datetime(step["deadline"]).strftime("%Y-%m-%d %H:%M:%S"),
        "client_order_ref": step["task_order_ref"],
    }
    sale_order_id = self._create(sale_order_model, sale_order_vals)
    sale_order_line_model.create(
        vals_list=[{
            "order_id": sale_order_id,
            "product_id": self._get_product_id("P674891CF17-SPP-001"),
            "product_uom_qty": step["demand"],
            "price_unit": step["price"],
        }]
    )
    sale_order_model.action_confirm(ids=[sale_order_id])

def _create_purchase_orders(self):
    PO, POLine = M("purchase.order"), M("purchase.order.line")
    for step in SOLVER_PLAN["purchases"]:
        # Resolve vendor, product, origin string, and planned date from the step.
        vendor_id = self._get_partner_id(step["vendor_ref"])
        product_id = self._get_product_id(step["product_code"])
        origin = self._format_po_origin(step["origin_customer_refs"])
        date_planned = self._scenario_datetime(step["planned_arrival_days"]).strftime("%Y-%m-%d %H:%M:%S")

        po_id = self._create(PO, {"partner_id": vendor_id, "origin": origin})
        POLine.create(
            vals_list=[{
                "order_id": po_id,
                "product_id": product_id,
                "product_qty": step["quantity"],
                "price_unit": step["unit_cost"],
                "date_planned": date_planned,
            }]
        )
        PO.button_confirm(ids=[po_id])
\end{Verbatim}

\section{Scoring and Verification}\label{app:scoring}

\subsection{Verifier Protocol}\label{e.1-verifier-protocol}

The verifier is a terminal-state program. \texttt{tests/test.sh} waits for the seeded Odoo to come up, starts \texttt{tests/checks.py} as a coprocess, and feeds it named check invocations of the form \texttt{check\_name\ arg1\ arg2\ ...} over stdin. Each check returns \texttt{PASS}, \texttt{FAIL}, or \texttt{NA} on stdout. The shell aggregates results, computes the reward, and writes these artifacts under \texttt{/logs/verifier}:

\begin{longtable}[]{@{}
  >{\raggedright\arraybackslash}p{0.4400\linewidth}
  >{\raggedright\arraybackslash}p{0.4400\linewidth}@{}}
\toprule\noalign{}
Artifact & Role \\
\midrule\noalign{}
\endhead
\bottomrule\noalign{}
\endlastfoot
\texttt{rule\_results.tsv} & Per-rule outcome with dimension, expression, and \texttt{PASS}/\texttt{FAIL}/\texttt{NA} status \\
\texttt{spend.json} & Realized spend, oracle spend, delta, and decay-based spend score \\
\texttt{optimality.json} & Primary objective value, oracle value, decay parameters, and final optimality score \\
\texttt{reward.txt} & Final reward as a single float \\
\texttt{reward.json} & Aggregated payload with overall score, per-dimension counts, per-rule status, and embedded \texttt{spend.json} and \texttt{optimality.json} \\
\texttt{checks.log} & Coprocess execution log \\
\end{longtable}

The reward grades the terminal Odoo state and ignores the agent trajectory. The same checks grade coding, browser, and computer-use runs.

Each check returns one of three statuses. \texttt{PASS} means the rule applies and the terminal ERP state satisfies it. \texttt{FAIL} means the rule applies and the state violates it. \texttt{NA} means the rule does not apply to this task or to the current terminal state --- for example an invoicing rule on a task with no invoice policy, or \texttt{forbidden\_finished\_po\_absent} on a task that allows buy routes. \texttt{PASS} and \texttt{FAIL} enter the per-dimension counters used by the reward formula in Section E.2; \texttt{NA} does not. The rule density numbers in Section E.6 reflect only checks that actually contributed to the score on a given task.

\subsection{Reward Formula}\label{e.2-reward-formula}

The verifier maintains three dimensions: \texttt{constraint}, \texttt{hygiene}, and \texttt{optimality}. For each rule, the shell increments a per-dimension \texttt{total} counter when the rule returns \texttt{PASS} or \texttt{FAIL} and a separate \texttt{earned} counter only on \texttt{PASS}. \texttt{NA} results stay out of both counters. Let

\[ c = 100 \cdot \frac{\text{constraint earned}}{\text{constraint total}}, \quad h = 100 \cdot \frac{\text{hygiene earned}}{\text{hygiene total}}, \quad o = \texttt{optimality\_score} \]

The final reward is

\[
R =
\begin{cases}
0 & \text{if a hard-zero gate fires (Section E.5),} \\
0.25\,c & \text{else if } c < 100, \\
0.25\,c + 0.60\,o + 0.15\,h & \text{otherwise.}
\end{cases}
\]

Constraint failures collapse the reward to the constraint slice alone. Optimality and hygiene only contribute once every applicable constraint passes.

\subsection{Constraint and Hygiene Checks}\label{e.3-constraint-and-hygiene-checks}

Constraint checks verify that the terminal ERP state encodes a feasible business plan. Hygiene checks verify documentation and isolation. Both dimensions reduce to a fixed catalog of named checks; each task's \texttt{tests/test.sh} instantiates the subset that applies to that scenario, with sampled arguments baked in. The tables below list the check functions and show the literal invocations as they appear in a task verifier.

\textbf{Constraint checks.}

\begin{longtable}[]{@{}
  >{\raggedright\arraybackslash}p{0.2933\linewidth}
  >{\raggedright\arraybackslash}p{0.2933\linewidth}
  >{\raggedright\arraybackslash}p{0.2933\linewidth}@{}}
\toprule\noalign{}
Check & Verifies & Example invocation \\
\midrule\noalign{}
\endhead
\bottomrule\noalign{}
\endlastfoot
\texttt{demand\_coverage} & Confirmed SO covers the required quantity for a customer/product & \texttt{demand\_coverage\ r674891CF17\_c01\ P674891CF17-SPP-001\ 4\ gte} \\
\texttt{deadline\_fulfillment} & Stock + linked PO/MO deliveries cover demand by the customer deadline & \texttt{deadline\_fulfillment\ r674891CF17\_c01\ P674891CF17-SPP-001\ 6} \\
\texttt{list\_price} & SO line price equals the catalog list price & \texttt{list\_price\ r674891CF17\_c01\ P674891CF17-SPP-001\ 442.85} \\
\texttt{sale\_revenue} & Confirmed SO revenue matches the retained target & \texttt{sale\_revenue\ r674891CF17\_c01\ P674891CF17-SPP-001\ 1771.4} \\
\texttt{budget\_compliance} & SO total stays within the customer pretax budget cap & \texttt{budget\_compliance\ r674891CF17\_c01\ P674891CF17-SPP-001\ 1921.97} \\
\texttt{new\_spend\_margin\_policy} & Portfolio-level new-spend margin clears the policy threshold & \texttt{new\_spend\_margin\_policy\ 0.287} \\
\texttt{supply\_timing\_feasible} & All confirmed supply orders arrive in time to cover linked demand & \texttt{supply\_timing\_feasible} \\
\texttt{supply\_coverage} & Stock plus confirmed supply meets total accepted demand & \texttt{supply\_coverage\ P674891CF17-SPP-001\ 32\ 27.0} \\
\texttt{po\_delivery\_schedule\_compliance} & PO planned dates match the vendor lead time on file & \texttt{po\_delivery\_schedule\_compliance} \\
\texttt{po\_min\_qty\_compliance} & PO quantity stays at or above the vendor minimum order quantity & \texttt{po\_min\_qty\_compliance} \\
\texttt{po\_price\_tier\_compliance} & PO line price matches the vendor offer for that quantity tier & \texttt{po\_price\_tier\_compliance} \\
\texttt{po\_consolidation\_compliance} & One PO per vendor/product (no fragmented purchases against the same offer) & \texttt{po\_consolidation\_compliance} \\
\texttt{finished\_vendor\_max\_qty\_compliance} & Cumulative purchases respect each vendor's maximum order quantity & \texttt{finished\_vendor\_max\_qty\_compliance} \\
\texttt{finished\_stock\_capacity\_compliance} & Stock allocations cannot exceed available on-hand finished stock & \texttt{finished\_stock\_capacity\_compliance} \\
\texttt{mo\_schedule\_compliance} & MO start/finish dates respect workcenter routing and downstream SO deadlines & \texttt{mo\_schedule\_compliance\ P5FE3FECD06-SAF-DCE-009} \\
\texttt{mo\_component\_feasibility} & MOs reserve enough components from stock or component POs & \texttt{mo\_component\_feasibility\ P5FE3FECD06-SAF-DCE-009} \\
\texttt{assembly\_capacity\_compliance} & Scheduled workcenter minutes stay below per-station capacity & \texttt{assembly\_capacity\_compliance} \\
\texttt{forbidden\_finished\_po\_absent} & Manufacture-only-by-policy tasks have no finished-goods POs & \texttt{forbidden\_finished\_po\_absent} \\
\texttt{forbidden\_finished\_mo\_absent} & Buy-only tasks have no finished-goods MOs & \texttt{forbidden\_finished\_mo\_absent} \\
\texttt{task\_state\_transitions\_completed} & Screened tasks confirm exactly the accepted orders and cancel the rest & \texttt{task\_state\_transitions\_completed} \\
\texttt{seeded\_order\_confirmed} / \texttt{seeded\_order\_reused} / \texttt{seeded\_order\_cancelled} & Per-order acceptance outcome on each preloaded draft & \texttt{seeded\_order\_cancelled\ r47CECE4885\_o03\ r47CECE4885\_c03\ P47CECE4885-HVC-SHP-002\ 21} \\
\texttt{prompt\_request\_not\_confirmed} & Prompt-only requests that should be rejected stay unconfirmed & \texttt{prompt\_request\_not\_confirmed\ r47CECE4885\_c01\ P47CECE4885-HVC-SHP-002} \\
\texttt{regular\_invoice\_amount\_matches\_policy} & Regular invoice total matches the order amount net of any deposit & \texttt{regular\_invoice\_amount\_matches\_policy\ r47CECE4885\_o09\ r47CECE4885\_c09\ prompt\_only\ P47CECE4885-HVC-SHP-002\ 88599.24} \\
\texttt{downpayment\_invoice\_amount\_matches\_policy} & Deposit invoice equals the fixed-amount or percentage policy slice & \texttt{downpayment\_invoice\_amount\_matches\_policy\ r47CECE4885\_o09\ r47CECE4885\_c09\ prompt\_only\ P47CECE4885-HVC-SHP-002\ 35439.70} \\
\end{longtable}

\textbf{Hygiene checks.}

\begin{longtable}[]{@{}
  >{\raggedright\arraybackslash}p{0.2933\linewidth}
  >{\raggedright\arraybackslash}p{0.2933\linewidth}
  >{\raggedright\arraybackslash}p{0.2933\linewidth}@{}}
\toprule\noalign{}
Check & Verifies & Example invocation \\
\midrule\noalign{}
\endhead
\bottomrule\noalign{}
\endlastfoot
\texttt{so\_confirmed} & A confirmed SO exists for the customer/product & \texttt{so\_confirmed\ r674891CF17\_c01\ P674891CF17-SPP-001} \\
\texttt{po\_origin\_traceability} & Each PO line lists the SOs it serves in the \texttt{origin} field, with feasible flow & \texttt{po\_origin\_traceability\ P674891CF17-SPP-001\ finished} \\
\texttt{mrp\_origin\_traceability} & Each MO lists the upstream SO or parent MO it serves & \texttt{mrp\_origin\_traceability\ P5FE3FECD06-SAF-DCE-009} \\
\texttt{retained\_order\_has\_expected\_payment\_term} & Retained SOs carry the policy payment term (e.g., \texttt{net\_30}) & \texttt{retained\_order\_has\_expected\_payment\_term\ r47CECE4885\_o09\ r47CECE4885\_c09\ prompt\_only\ P47CECE4885-HVC-SHP-002} \\
\texttt{linked\_posted\_invoices\_match\_order\_partner} & Posted invoices belong to the SO partner & \texttt{linked\_posted\_invoices\_match\_order\_partner\ r47CECE4885\_o09\ r47CECE4885\_c09\ prompt\_only\ P47CECE4885-HVC-SHP-002} \\
\texttt{linked\_posted\_invoices\_are\_tax\_free} & Posted invoices have zero tax, matching the seeded fiscal setup & \texttt{linked\_posted\_invoices\_are\_tax\_free\ r47CECE4885\_o09\ r47CECE4885\_c09\ prompt\_only\ P47CECE4885-HVC-SHP-002} \\
\texttt{rejected\_order\_not\_invoiced} & Rejected requests have no posted invoices & \texttt{rejected\_order\_not\_invoiced\ r47CECE4885\_o01\ r47CECE4885\_c01\ prompt\_only\ P47CECE4885-HVC-SHP-002} \\
\texttt{adjacent\_data\_untouched} & The 80--200 unrelated customers, vendors, and products in the seeded ERP have no new or modified records & \texttt{adjacent\_data\_untouched} \\
\end{longtable}

\texttt{repair\_state\_compliance} and \texttt{has\_relevant\_activity} also drive hard-zero gates (Section E.5). \texttt{task\_state\_transitions\_completed} is a scored constraint check, but it does not by itself zero the whole reward.

\subsection{Optimality Calculation}\label{e.4-optimality-calculation}

The verifier recomputes the realized primary objective from terminal Odoo records and compares it with the solver-certified oracle value. The four objectives that have an optimization phase share the same exponential decay function

\[
\texttt{score}(a,e) =
\begin{cases}
100 & \text{if } a \le e + \tau, \\
100 \cdot \exp\!\left(-k \cdot \dfrac{a-e}{\max(e,1)}\right) & \text{otherwise.}
\end{cases}
\]

where $a$ is the realized primary metric, $e$ is the oracle value, $\tau$ is a small tolerance, and $k$ controls how quickly the score decays as the agent overshoots the oracle.

\begin{longtable}[]{@{}
  >{\raggedright\arraybackslash}p{0.1760\linewidth}
  >{\raggedright\arraybackslash}p{0.1760\linewidth}
  >{\raggedleft\arraybackslash}p{0.1760\linewidth}
  >{\raggedleft\arraybackslash}p{0.1760\linewidth}
  >{\raggedright\arraybackslash}p{0.1760\linewidth}@{}}
\toprule\noalign{}

Objective & Primary metric ($a$) & Tolerance ($\tau$) & Decay ($k$) & Secondary metric \\

\midrule\noalign{}
\endhead
\bottomrule\noalign{}
\endlastfoot
\texttt{min\_new\_spend} & Total realized new spend (finished POs + component POs + manufacturing cost from \texttt{WORKCENTER\_OPTION\_COSTS}) & 0.25 & 5.0 & none \\
\texttt{vendor\_consolidation} & Number of distinct relevant vendors used & 0 & 2.0 & spend \\
\texttt{capacity\_preservation} & Total scheduled workcenter minutes used & 0.01 & 5.0 & spend \\
\texttt{repair\_plan} & (L\_1) penalty between realized and seeded baseline supply mix (per offer and per \texttt{(product,\ workcenter)} routing) & 0 & 2.0 & spend \\
\texttt{constraint\_only} & n/a & n/a & n/a & none (optimality fixed at 100) \\
\end{longtable}

The realized spend used by the secondary score is computed from confirmed POs by resolving each line against the vendor offer tier on file (\texttt{\_resolve\_tier\_price}) so the verifier never trusts the agent's price field. Manufacturing cost is added by multiplying realized MO quantity by the per-route unit cost in \texttt{WORKCENTER\_OPTION\_COSTS}. The result is compared against \texttt{EXPECTED\_SPEND} and decayed with 
($k = 5.0$) and ($\tau = 0.25$)
. The repair penalty is the sum of absolute differences between realized and baseline quantities at offer-key and \texttt{(product,\ workcenter)} granularity, so the optimality score rewards plans that stay as close to the seeded baseline as the disruption allows.

For objectives with a secondary spend metric, the verifier combines the primary score (p) and the spend score (s) using a fixed lexicographic blend with (w = \texttt{SECONDARY\_SPEND\_BAND\_WEIGHT} = 0.10):

\[
\texttt{optimality\_score}(p,s) =
\begin{cases}
(1-w)\cdot 100 + w\cdot s & \text{if } p \ge 100, \\
(1-w)\cdot p & \text{if } p < 100.
\end{cases}
\]

The agent has to clear the primary objective (vendor count, scheduled minutes, or repair distance) before spend can move the score above 90. Any primary shortfall caps optimality at (0.9 p), so spend can never compensate for a primary regression.

\subsection{Hard-Zero Gates}\label{e.5-hard-zero-gates}

Two structural conditions collapse the entire reward to 0 even when individual constraint checks pass.

\begin{longtable}[]{@{}
  >{\raggedright\arraybackslash}p{0.2933\linewidth}
  >{\raggedright\arraybackslash}p{0.2933\linewidth}
  >{\raggedright\arraybackslash}p{0.2933\linewidth}@{}}
\toprule\noalign{}
Gate & Trigger & Affected tasks \\
\midrule\noalign{}
\endhead
\bottomrule\noalign{}
\endlastfoot
\texttt{PARTIAL\_ACCEPTANCE\_ACTIVITY\_OK} & \texttt{has\_relevant\_activity} returns \texttt{FAIL} because the agent left every task-relevant record untouched & Screened tasks (intake mode = \texttt{screen\_orders\_by\_policy}) \\
\texttt{REPAIR\_STATE\_OK} & \texttt{repair\_state\_compliance} returns \texttt{FAIL} because the seeded broken plan is still present and unrepaired & Repair-plan tasks \\
\end{longtable}

Both gates feed \texttt{HARD\_SCORE\_GATE\_OK} in \texttt{tests/test.sh}. When that gate is 0, the final-score branch in the awk formula returns \texttt{0.00} regardless of constraint, optimality, or hygiene partial credit. The gates exist because partial-acceptance tasks reward an explicit triage decision (a no-op terminal state never demonstrates one) and repair tasks measure a deviation from the seeded baseline (an unmodified state has zero deviation but no actual repair). Without the gates, both task types would reward inactivity.

\subsection{Rule Density}\label{e.6-rule-density}

\begin{longtable}[]{@{}lrrrl@{}}
\toprule\noalign{}
Difficulty & Tasks & Mean scored checks & Median & Range \\
\midrule\noalign{}
\endhead
\bottomrule\noalign{}
\endlastfoot
Easy & 50 & 55.76 & 63 & 38 to 66 \\
Medium & 166 & 73.01 & 76 & 29 to 146 \\
Hard & 84 & 122.76 & 105 & 66 to 208 \\
Overall & 300 & 84.07 & 78 & 29 to 208 \\
\end{longtable}

Hard tasks combine more dimensions of the ERP state, so each scenario expands into more independent checks. The denominator counts only applicable checks (\texttt{PASS} plus \texttt{FAIL}); inapplicable checks return \texttt{NA} and stay out of the rule density numbers.

\subsection{Rule Families}\label{e.7-rule-families}

\begin{longtable}[]{@{}
  >{\raggedright\arraybackslash}p{0.2933\linewidth}
  >{\raggedright\arraybackslash}p{0.2933\linewidth}
  >{\raggedleft\arraybackslash}p{0.2933\linewidth}@{}}
\toprule\noalign{}
Rule family & Representative checks & Scored checks across tasks \\
\midrule\noalign{}
\endhead
\bottomrule\noalign{}
\endlastfoot
Demand and sales & \texttt{demand\_coverage}, \texttt{deadline\_fulfillment}, \texttt{list\_price}, \texttt{sale\_revenue}, \texttt{budget\_compliance} & 15,933 \\
Supply and procurement & \texttt{supply\_timing\_feasible}, \texttt{supply\_coverage}, \texttt{po\_delivery\_schedule\_compliance}, \texttt{po\_min\_qty\_compliance}, \texttt{po\_price\_tier\_compliance}, \texttt{po\_consolidation\_compliance}, \texttt{finished\_vendor\_max\_qty\_compliance}, \texttt{finished\_stock\_capacity\_compliance}, \texttt{new\_spend\_margin\_policy} & 2,840 \\
Invoicing and payment & \texttt{regular\_invoice\_amount\_matches\_policy}, \texttt{downpayment\_invoice\_amount\_matches\_policy}, \texttt{linked\_posted\_invoices\_match\_order\_partner}, \texttt{linked\_posted\_invoices\_are\_tax\_free}, \texttt{retained\_order\_has\_expected\_payment\_term}, \texttt{rejected\_order\_not\_invoiced} & 2,132 \\
Traceability and hygiene & \texttt{so\_confirmed}, \texttt{po\_origin\_traceability}, \texttt{mrp\_origin\_traceability}, \texttt{adjacent\_data\_untouched} & 1,935 \\
Manufacturing and capacity & \texttt{mo\_schedule\_compliance}, \texttt{mo\_component\_feasibility}, \texttt{assembly\_capacity\_compliance}, \texttt{forbidden\_finished\_mo\_absent} & 1,404 \\
State and seeded orders & \texttt{task\_state\_transitions\_completed}, \texttt{seeded\_order\_confirmed}, \texttt{seeded\_order\_reused}, \texttt{seeded\_order\_cancelled}, \texttt{prompt\_request\_not\_confirmed}, \texttt{repair\_state\_compliance}, \texttt{has\_relevant\_activity} & 976 \\
\end{longtable}

Demand and sales dominates because every task begins with customer demand, and each customer/product pair drives separate \texttt{demand\_coverage}, \texttt{deadline\_fulfillment}, \texttt{list\_price}, \texttt{sale\_revenue}, and \texttt{budget\_compliance} checks. The other families enter only when the sampled pattern requires manufacturing, screening, invoicing, or repair logic.

\section{Benchmark Quality Control}\label{app:quality-control}

\subsection{Oracle Replay}\label{f.1-oracle-replay}

Oracle replay executes \texttt{solution/solve.sh} against each seeded task environment. That script calls \texttt{solution/solver.py}, which writes the plan encoded in \texttt{solution/optimal\_plan.json} into Odoo through the JSON-2 API. The verifier then grades the terminal Odoo state through the same \texttt{tests/test.sh} path used for agent trials.

On the 300-task corpus used for this draft, oracle replay reaches full reward on 300 of 300 tasks.

\subsection{No-op Baseline}\label{f.2-no-op-baseline}

The no-op baseline starts the same seeded task environment and runs \texttt{tests/test.sh} without invoking an agent or oracle writer. The expected terminal state is the untouched setup state. This check catches tasks that start solved and verifier rules that accidentally accept missing business activity.

On the 300-task corpus used for this draft, the no-op baseline receives zero reward on 300 of 300 tasks.

\subsection{Reward-Hacking Canary}\label{f.3-reward-hacking-canary}

The canary value in ERP-Bench is realized spend. The agent has write access to the PO line \texttt{price\_unit} field, so a naive verifier could be tricked into rewarding an agent that simply writes a small number into that field. The verifier defends spend in two ways.

The optimality dimension recomputes spend from the authoritative vendor offers rather than reading the agent-written \texttt{price\_unit}. Each PO line is resolved through \texttt{\_resolve\_tier\_price(vendor,\ product,\ qty)}, which returns the canonical price for the matching min-quantity tier in \texttt{SCENARIO\_SUPPLIER\_OFFERS}. The realized spend is then \texttt{qty\ *\ tier\_price}, independent of what the agent typed.

The constraint dimension separately enforces that the agent's \texttt{price\_unit} actually matches the tier price. \texttt{po\_price\_tier\_compliance} fails if any relevant PO line is outside a small tolerance around the canonical price:

\begin{Verbatim}[breaklines=true,breakanywhere=true,fontsize=\footnotesize]
def po_price_tier_compliance():
    for row in _relevant_po_rows():
        tp = _resolve_tier_price(row["vendor_id"], row["product_id"], row["qty"])
        if tp is None or abs(row["price_unit"] - tp) > max(0.02, tp * 1e-4):
            return False
    return True
\end{Verbatim}

A failed \texttt{po\_price\_tier\_compliance} clips the score through the constraint gate (Section E.2), so an agent that writes a wrong price loses constraint credit even if optimality is unaffected.

Empirically, the canary scan found no perfect-score trial whose realized primary objective beat the oracle value on the 300-task corpus.

\subsection{Task Reviewer Agent}\label{f.4-task-reviewer-agent}

The release ships an LLM reviewer agent that audits generated tasks before they merge into the corpus. The agent enforces an alignment chain that crosses every artifact in a task bundle:

\begin{Verbatim}[breaklines=true,breakanywhere=true,fontsize=\footnotesize]
solver <-> instruction <-> data loader <-> verifier
\end{Verbatim}

For each in-scope task, the agent reads \texttt{task.toml}, \texttt{instruction.md}, \texttt{environment/setup\_scenario.py}, \texttt{environment/scenario\_data.json}, \texttt{solution/optimal\_plan.json}, \texttt{solution/solver.py}, and \texttt{tests/checks.py}, and looks for mismatches such as instruction text that promises constraints the verifier never checks, oracle plans that violate constraints the instruction declares, or scenario data that the solver assumes but the loader never seeds. It also validates the realism of model and field usage against the upstream Odoo 19 source and documentation, since the verifier ultimately reads Odoo records and a check that depends on a field with the wrong semantics is a silent benchmark bug.

Outputs follow a fixed format: severity-ordered findings (Critical / Major / Minor) with the impacted task, the exact mismatch in the alignment chain, evidence, the likely impact on scoring, solvability, or realism, and a concrete fix recommendation in source-of-truth files (\texttt{erp\_bench/}, \texttt{schemas/}, or \texttt{tests/}) rather than manual edits under \texttt{tasks/}. Each review ends with open questions, a per-task pass/fail summary, and the validation evidence (\texttt{uv\ run\ pytest}, plus \texttt{harbor\ run\ -p\ \textless{}scope\textgreater{}\ -a\ nop} and \texttt{harbor\ run\ -p\ \textless{}scope\textgreater{}\ -a\ oracle} invocations).

Default behavior is review-first: the agent reports findings without modifying code unless the user explicitly asks for fixes. This keeps the reviewer auditable --- every change to the corpus is traceable to either a deterministic generator update or a reviewed manual fix.

\section{Evaluation Protocol}\label{app:harness}

\subsection{Harness Overview}\label{g.1-harness-overview}

The three harnesses share the same agent CLI and the same Odoo task environment. They differ only in which tools the agent is allowed to call and how those tools observe the ERP. The shared agent CLI is \texttt{pi}, an open-source coding agent from \href{https://github.com/agentic-labs/pi-mono}{\texttt{agentic-labs/pi-mono}}. ERP-Bench wraps \texttt{pi} in three thin Python adapters under \texttt{agents/} in the release: \texttt{pi.py} for the coding harness, \texttt{pi\_browser\_use.py} for the browser harness, and \texttt{pi\_computer\_use.py} for the computer-use harness. The browser and computer-use adapters add a single \texttt{-\/-extension} flag that loads a TypeScript module from \texttt{pi-mono/extensions/} and replaces the default file/shell tool surface with one UI tool.

\begin{longtable}[]{@{}
  >{\raggedright\arraybackslash}p{0.2200\linewidth}
  >{\raggedright\arraybackslash}p{0.2200\linewidth}
  >{\raggedright\arraybackslash}p{0.2200\linewidth}
  >{\raggedright\arraybackslash}p{0.2200\linewidth}@{}}
\toprule\noalign{}
Harness & Adapter & Pi extension & Tool surface visible to the agent \\
\midrule\noalign{}
\endhead
\bottomrule\noalign{}
\endlastfoot
Coding & \texttt{agents/pi.py} & none (Pi defaults) & \texttt{read}, \texttt{write}, \texttt{edit}, \texttt{bash}, \texttt{grep}, \texttt{find}, \texttt{ls} \\
Browser & \texttt{agents/pi\_browser\_use.py} & \texttt{pi-mono/extensions/browser-use/index.ts} & one \texttt{browser} tool that batches up to 7 accessibility-tree actions per call \\
Computer-use & \texttt{agents/pi\_computer\_use.py} & \texttt{pi-mono/extensions/computer-use/index.ts} & one \texttt{computer} tool that batches up to 16 mouse/keyboard/screenshot actions per call \\
\end{longtable}

The instruction body and verifier are identical across harnesses, enabling end-to-end comparisons of agent systems with different interaction interfaces and tool access.

\subsection{Pi Coding Agent}\label{g.2-pi-coding-agent}

The base adapter \texttt{agents/pi.py} installs \texttt{pi} from a pinned \texttt{pi-mono} git tag, then runs each trial with

\begin{Verbatim}[breaklines=true,breakanywhere=true,fontsize=\footnotesize]
pi --print --mode json --no-session --provider <p> --model <m> [--thinking <level>] [--max-turns <N>] '<instruction>'
\end{Verbatim}

\texttt{-\/-print\ -\/-mode\ json} makes \texttt{pi} emit one JSON event per line on stdout (including \texttt{message\_end} events with token and cost usage), which the adapter tees to \texttt{/logs/agent/pi.txt} for later parsing into the trial's \texttt{context} payload. \texttt{-\/-no-session} starts a fresh conversation per trial with no resume state from prior trials; the agent itself still loops across many tool-call turns inside that conversation until it finishes or hits the turn budget. The provider and model name come from the harness configuration; the adapter forwards the matching API key environment variables (\texttt{OPENAI\_API\_KEY}, \texttt{ANTHROPIC\_API\_KEY}, etc.) into the container.

Pi's default system prompt (built in \texttt{pi-mono/packages/coding-agent/src/core/system-prompt.ts}) declares the agent as ``an expert coding assistant operating inside pi'', lists the available tools as a bulleted snippet, attaches a short list of tool-aware guidelines (for example ``Prefer grep/find/ls tools over bash for file exploration''), appends any \texttt{\#\ Project\ Context} files, and ends with the current date and working directory. Each tool ships its own JSON-Schema parameter definition; the seven default coding tools are \texttt{read}, \texttt{write}, \texttt{edit}, \texttt{bash}, \texttt{grep}, \texttt{find}, and \texttt{ls}. The ERP-Bench \texttt{instruction.md} (Appendix D.2) becomes the first user message; Pi loops between assistant tool calls and tool results until the agent finishes or the harness-imposed turn budget is exhausted.

Inside the task container, the coding harness reaches Odoo through the Odoo JSON-RPC API on \texttt{http://127.0.0.1:8069}, called from \texttt{bash} and \texttt{python3} scripts the agent writes through \texttt{write} and \texttt{edit}. Filesystem tools and shell access give the agent room to write helper scripts, but every business effect still has to flow through Odoo.

\subsection{Browser-Use Extension}\label{g.3-browser-use-extension}

The browser harness loads \texttt{pi-mono/extensions/browser-use/index.ts} through \texttt{pi\ -\/-extension}. The extension launches a headless Chromium through Playwright on the first tool call and exposes a single tool, \texttt{browser}, whose parameter schema accepts an \texttt{actions} list of up to seven entries. The action vocabulary is

\begin{Verbatim}[breaklines=true,breakanywhere=true,fontsize=\footnotesize]
goto(url) | snapshot | click(ref) | fill(ref, text) | press(key, ref?) |
scroll(ref?, direction?, amount?) | wait(ms?, loadState?)
\end{Verbatim}

Refs are opaque accessibility-tree handles from Playwright's \texttt{aria-ref=} locator. Every successful \texttt{browser} call returns a fresh accessibility-tree snapshot, the page URL, and the page title in a single text block; the agent never receives a screenshot.

On startup the extension calls \texttt{setActiveTools({[}"browser"{]})}, which masks Pi's default coding tools so the model literally cannot call \texttt{read}, \texttt{bash}, \texttt{edit}, or any other tool --- any non-\texttt{browser} call is intercepted and rejected with \texttt{"browser-use\ enforces\ browser-only\ mode"}. The extension also injects a tool snippet and a guideline list into the system prompt (declared as \texttt{DEFAULT\_BROWSER\_PROMPT\_SNIPPET} and \texttt{DEFAULT\_BROWSER\_GUIDELINES}) covering ref handling, batching, the no-screenshots rule, and recovery on click errors that nevertheless changed the page. To control prompt-cache costs on Anthropic, the extension intercepts outgoing provider requests and moves the \texttt{cache\_control} marker to the block immediately preceding the latest browser tool result, then collapses earlier accessibility trees to short summaries through a \texttt{context} event hook.

\subsection{Computer-Use Extension}\label{g.4-computer-use-extension}

The computer-use harness loads \texttt{pi-mono/extensions/computer-use/index.ts} and exposes a single tool, \texttt{computer}, with the action vocabulary

\begin{Verbatim}[breaklines=true,breakanywhere=true,fontsize=\footnotesize]
screenshot | click(x, y, button?, modifiers?) | double_click(x, y) | drag(path[]) |
move(x, y) | scroll(x, y, scrollX, scrollY) | type(text) | keypress(keys[]) |
mouse_down(x, y, button?) | mouse_up(x, y, button?) | hold_key(key, duration?) | wait(seconds?)
\end{Verbatim}

The action batch is capped at 16 entries per call, and every successful call returns a single screenshot of the resulting display. Coordinates are pixels from the top-left corner of the display.

Driving the desktop is delegated to standard Linux tools rather than Playwright. \texttt{agents/pi\_computer\_use.py} installs \texttt{xvfb}, \texttt{xdotool}, \texttt{scrot}, \texttt{fluxbox}, and \texttt{dbus-x11} at container build time, and a \texttt{PRE\_RUN\_COMMAND} script starts an \texttt{Xvfb} virtual display, a \texttt{fluxbox} window manager, and a Chromium process pointed at \texttt{http://127.0.0.1:8069/web/login} before the agent begins. The extension reads display geometry and timing from \texttt{PI\_COMPUTER\_USE\_DISPLAY*}, \texttt{PI\_COMPUTER\_USE\_ACTION\_DELAY\_MS}, \texttt{PI\_COMPUTER\_USE\_CLICK\_DELAY\_MS}, and \texttt{PI\_COMPUTER\_USE\_TYPING\_DELAY\_MS} (defaults: 1024$\times$768, 80 ms per action, 12 ms per character). Each tool call shells out to \texttt{xdotool} for mouse and keyboard events and to \texttt{scrot\ -\/-overwrite\ \textless{}path\textgreater{}} for screenshots; the resulting PNG is base64-encoded and returned as an \texttt{image} content block to the model.

As with the browser extension, the computer-use extension calls \texttt{setActiveTools({[}"computer"{]})} so all of Pi's default coding tools are unavailable and any other tool call is blocked with \texttt{"computer-use\ enforces\ computer-only\ mode"}. The extension also injects its own system-prompt snippet and guideline list (\texttt{DEFAULT\_COMPUTER\_PROMPT\_SNIPPET}, \texttt{DEFAULT\_COMPUTER\_GUIDELINES}) covering screenshot batching, the always-included final screenshot, and the rule to take a fresh screenshot before acting on coordinates that may have shifted.

\subsection{Shared Trial Settings}\label{g.5-shared-trial-settings}

Each complete model-harness cell attempts 300 tasks with five trials per task. Each trial receives a one-hour timeout and a 400-turn budget. Coding and browser runs evaluate four model variants. Computer-use runs evaluate GPT-5.5, Claude Opus 4.7, GLM-5v, and Kimi K2.5. GLM-5v replaces GLM-5.1 in computer-use runs, since the computer-use harness requires vision input.

The GLM-5v and Kimi K2.5 computer-use cells were halted early after zero-point partial runs. Their completed attempts remain useful for reporting failure behavior. Aggregate complete-cell statistics use the ten full 1,500-attempt cells unless explicitly stated otherwise.

\subsection{Recomputation Inputs}\label{g.6-recomputation-inputs}

The metric recomputation in this appendix reads \texttt{reward.json} from each per-trial output directory. Missing, malformed, or unparsable \texttt{reward.json} files count as zero-score failed attempts. Parseable reward files provide the score even when the per-trial \texttt{result.json} records an exception. Per-cell totals (attempted trials, parseable rewards, run exceptions) appear in Appendix H.1. The Kimi coding numbers come from the HTML comparison report rather than the raw run directory, since one earlier run directory was incomplete and the user-facing report is the canonical source for that cell.

\section{Full Quantitative Results}\label{app:full-results}

\subsection{Headline Results}\label{h.1-headline-results}

{\footnotesize
\setlength{\tabcolsep}{2pt}
\begin{longtable}[]{@{}
  >{\raggedright\arraybackslash}p{0.0800\linewidth}
  >{\raggedright\arraybackslash}p{0.0800\linewidth}
  >{\raggedleft\arraybackslash}p{0.0800\linewidth}
  >{\raggedleft\arraybackslash}p{0.0800\linewidth}
  >{\raggedleft\arraybackslash}p{0.0800\linewidth}
  >{\raggedleft\arraybackslash}p{0.0800\linewidth}
  >{\raggedleft\arraybackslash}p{0.0800\linewidth}
  >{\raggedleft\arraybackslash}p{0.0800\linewidth}
  >{\raggedleft\arraybackslash}p{0.0800\linewidth}
  >{\raggedleft\arraybackslash}p{0.0800\linewidth}
  >{\raggedleft\arraybackslash}p{0.0800\linewidth}@{}}
\toprule\noalign{}
Harness & Model & Attempts & Parseable rewards & Run exceptions & Pass@1 & Pass@5 & Avg score & Constraints clean & Perfect score & Feasible optimality \\
\midrule\noalign{}
\endhead
\bottomrule\noalign{}
\endlastfoot
Coding & GPT-5.5 & 1,500 & 1,455 & 49 & 43.4 & 73.0 & 63.6 & 59.2 & 36.5 & 90.8 \\
Coding & Claude Opus 4.7 & 1,500 & 1,475 & 26 & 30.8 & 60.7 & 52.2 & 42.5 & 27.6 & 94.7 \\
Coding & GLM-5.1 & 1,500 & 1,476 & 27 & 35.8 & 63.3 & 53.7 & 45.7 & 30.6 & 95.6 \\
Coding & Kimi K2.5 & 1,500 & 1,410 & 108 & 9.1 & 20.3 & 18.5 & 10.4 & 8.9 & 92.8 \\
Browser & GPT-5.5 & 1,500 & 1,346 & 156 & 9.7 & 23.7 & 16.8 & 12.7 & 9.5 & 90.3 \\
Browser & Claude Opus 4.7 & 1,500 & 1,435 & 69 & 28.8 & 44.3 & 47.8 & 40.1 & 27.3 & 86.8 \\
Browser & GLM-5.1 & 1,500 & 1,439 & 190 & 2.4 & 12.0 & 3.5 & 2.8 & 2.4 & 95.1 \\
Browser & Kimi K2.5 & 1,500 & 1,452 & 129 & 1.5 & 7.3 & 2.7 & 1.7 & 1.5 & 98.5 \\
Computer-use & GPT-5.5 & 1,500 & 1,444 & 57 & 7.9 & 16.7 & 20.0 & 9.5 & 7.7 & 93.4 \\
Computer-use & Claude Opus 4.7 & 1,500 & 1,426 & 76 & 23.7 & 38.3 & 40.2 & 34.0 & 22.3 & 88.3 \\
Computer-use & GLM-5v & 540 & 449 & 230 & 0.0 & 0.0 & 0.0 & 0.0 & 0.0 & NA \\
Computer-use & Kimi K2.5 & 619 & 526 & 209 & 0.0 & 0.0 & 0.1 & 0.0 & 0.0 & NA \\
\end{longtable}
}

The strongest coding cell is materially higher than the strongest browser or computer-use cell. UI-based cells also have more execution failures, visible in the missing or malformed reward files.

\subsection{Results by Harness}\label{h.2-results-by-harness}

This table includes all completed attempts, including the two halted computer-use cells.

\begin{longtable}[]{@{}
  >{\raggedright\arraybackslash}p{0.1467\linewidth}
  >{\raggedleft\arraybackslash}p{0.1467\linewidth}
  >{\raggedleft\arraybackslash}p{0.1467\linewidth}
  >{\raggedleft\arraybackslash}p{0.1467\linewidth}
  >{\raggedleft\arraybackslash}p{0.1467\linewidth}
  >{\raggedleft\arraybackslash}p{0.1467\linewidth}@{}}
\toprule\noalign{}
Harness & Attempts & Pass@1 & Average per-cell pass@5 & Constraints clean & Perfect score \\
\midrule\noalign{}
\endhead
\bottomrule\noalign{}
\endlastfoot
Coding & 6,000 & 29.8 & 54.3 & 39.5 & 25.9 \\
Browser & 6,000 & 10.6 & 21.8 & 14.3 & 10.2 \\
Computer-use & 4,159 & 11.4 & 13.8 & 15.7 & 10.8 \\
\end{longtable}

Coding agents solve more tasks under the same verifier. Browser and computer-use agents face higher interaction burden and execution fragility.

\subsection{Results by Difficulty}\label{h.3-results-by-difficulty}

This table includes all completed attempts, including the two halted computer-use cells.

\begin{longtable}[]{@{}
  >{\raggedright\arraybackslash}p{0.1467\linewidth}
  >{\raggedleft\arraybackslash}p{0.1467\linewidth}
  >{\raggedleft\arraybackslash}p{0.1467\linewidth}
  >{\raggedleft\arraybackslash}p{0.1467\linewidth}
  >{\raggedleft\arraybackslash}p{0.1467\linewidth}
  >{\raggedleft\arraybackslash}p{0.1467\linewidth}@{}}
\toprule\noalign{}
Difficulty & Attempts & Pass@1 & Constraints clean & Perfect score & Average score \\
\midrule\noalign{}
\endhead
\bottomrule\noalign{}
\endlastfoot
Easy & 2,689 & 35.6 & 41.5 & 34.5 & 45.6 \\
Medium & 8,949 & 17.8 & 25.1 & 15.5 & 29.7 \\
Hard & 4,521 & 7.8 & 11.4 & 6.6 & 19.9 \\
\end{longtable}

Pass rates, constraint satisfaction, perfect score, and average score all decline from easy to medium to hard.

\subsection{Task Parameters Predict Actions Needed For Resolution}\label{h.4-task-parameters-predict-actions-needed-for-resolution}

Figure 5. Task parameters also predict action burden among resolved trials. Correlations are reported separately by harness because coding, browser, and computer-use agents use different action surfaces.

\section{Failure Analysis}\label{app:failure-analysis}

\subsection{Failure Categories}\label{i.1-failure-categories}

ERP-Bench parseable failures fall into five broad categories:

\begin{enumerate}
\def\labelenumi{\arabic{enumi}.}
\item
  Demand and sales failures, where the agent fails to create, confirm, price, or cover the required sales orders.
\item
  Timing failures, where purchases, manufacturing orders, or deliveries arrive after the customer deadline.
\item
  Sourcing failures, where the agent violates vendor price, lead time, minimum quantity, maximum quantity, consolidation, screened-vendor, spend-floor, or margin policy.
\item
  Manufacturing failures, where the agent creates an infeasible production plan with incompatible components, workcenters, or capacities.
\item
  Hygiene and state failures, where the agent satisfies some business quantities but breaks traceability, confirmation state, invoicing, seeded-order handling, repair state, or adjacent-data preservation.
\end{enumerate}

The categories are not exclusive: one terminal ERP state can fail demand, timing, sourcing, and hygiene checks at once. The table below labels each parseable failed trial by the families of failed verifier rules in \texttt{rule\_results.tsv}. Category percentages use parseable failed trials as the denominator. We do not use optimality to explain these rows when feasibility fails. The optimality score is meaningful only after the terminal state satisfies the constraint and hygiene rules.

The rule-level coding counts use 5,047 raw attempts because the available Kimi K2.5 coding directory contains 547 attempts; Appendix H.1 reports the complete Kimi coding cell from the canonical HTML report.

\begin{longtable}[]{@{}
  >{\raggedright\arraybackslash}p{0.1100\linewidth}
  >{\raggedleft\arraybackslash}p{0.1100\linewidth}
  >{\raggedleft\arraybackslash}p{0.1100\linewidth}
  >{\raggedleft\arraybackslash}p{0.1100\linewidth}
  >{\raggedleft\arraybackslash}p{0.1100\linewidth}
  >{\raggedleft\arraybackslash}p{0.1100\linewidth}
  >{\raggedleft\arraybackslash}p{0.1100\linewidth}
  >{\raggedleft\arraybackslash}p{0.1100\linewidth}@{}}
\toprule\noalign{}
Harness & Raw attempts & Parseable failed trials & Demand/sales & Timing & Sourcing & Manufacturing & Hygiene/state \\
\midrule\noalign{}
\endhead
\bottomrule\noalign{}
\endlastfoot
Coding & 5,047 & 3,482 & 455 (13.1\%) & 2,399 (68.9\%) & 1,492 (42.8\%) & 1,217 (35.0\%) & 1,323 (38.0\%) \\
Browser & 6,000 & 5,062 & 3,341 (66.0\%) & 4,415 (87.2\%) & 4,509 (89.1\%) & 984 (19.4\%) & 3,972 (78.5\%) \\
Computer-use & 4,159 & 3,395 & 1,608 (47.4\%) & 3,073 (90.5\%) & 2,912 (85.8\%) & 717 (21.1\%) & 2,347 (69.1\%) \\
\end{longtable}

Coding failures are mostly plan-execution failures after the agent has enough information to construct the demand side. The common coding mistake is to create a plausible ERP object graph but schedule production before inputs can legally arrive, mutate confirmed procurements after discovering a planning error, or lose origin links while repairing the plan.

Browser failures more often begin with UI state ambiguity. The browser agent can reason about the right quantities but still leave a sales-order line blank, fail to confirm a quotation, or overwrite a procurement date after recognizing the selected vendor is late.

Computer-use failures have the same business-rule structure as browser failures, but the screenshot interface makes fine-grained form editing harder. The characteristic mistake is not a different ERP concept; it is committing a partially edited PO or SO because the agent cannot reliably tell whether the cell, datepicker, autocomplete, or source field accepted the intended value.

\subsection{Failure Evidence by Harness}\label{i.2-failure-evidence-by-harness}

The following examples are randomly sampled failed trajectories from the supplied raw directories using a fixed seed (\texttt{7}), filtered to parseable rewards with concrete failed verifier rules. Each example traces the verifier failure to the agent decision that caused it.

\subsubsection{Coding: Manufacturing Before Inputs Arrive}\label{coding-manufacturing-before-inputs-arrive}

Trial: GLM-5.1, \texttt{2126\_medium\_13\_single\_subassembl\_\_9Nkydd6} on \texttt{2126\_medium\_13\_single\_subassembly\_qualified\_workcenters}.

Verifier signal: \texttt{supply\_timing\_feasible}, \texttt{po\_delivery\_schedule\_compliance}, \texttt{mo\_schedule\_compliance}, \texttt{component\_stock\_capacity\_compliance}, \texttt{mo\_component\_feasibility}, and \texttt{po\_origin\_traceability} fail.

The agent correctly computed that it needed 64 new finished units, but it scheduled manufacturing at the start of the run and then force-received components through the API before respecting supplier lead times:

\begin{quote}
``The key steps are: 1. Create 9 Sales Orders 2. Create 1 MO for 64 finished products \ldots{} 3. Create 1 MO for 64 subassemblies \ldots{} 4. Create 8 POs for components \ldots{} 5. Confirm everything''
\end{quote}

\begin{quote}
``All POs confirmed. Now let me receive the goods and process the manufacturing orders''
\end{quote}

\begin{quote}
``All PO receipts validated. Now I can see the stock levels \ldots{} Now I need to process the manufacturing orders''
\end{quote}

The trace also shows the second mistake. After the first MOs became invalid, the agent cancelled and recreated the MOs, but kept the already-created component POs and tried to repair lineage after the fact:

\begin{quote}
``The POs for components were created with origins pointing to the cancelled MOs (WH/MO/00001 and WH/MO/00002). The active MOs are now WH/MO/00003 and WH/MO/00004. I need to update the PO origins.''
\end{quote}

What the agent should have done: compute the earliest component-availability date from existing component stock plus PO lead times, schedule the subassembly MO no earlier than that date, schedule the finished-product MO after the subassembly MO, and create confirmed component POs whose \texttt{origin} fields point to the active consuming MO before receiving or consuming anything. Once it cancelled the first MOs, it should have cancelled and recreated the dependent POs instead of mutating confirmed procurements.

\subsubsection{Browser: Proceeding After A Broken Sales-Order Row}\label{browser-proceeding-after-a-broken-sales-order-row}

Trial: GPT-5.5, \texttt{2123\_medium\_13\_single\_subassembl\_\_stwVQVA} on \texttt{2123\_medium\_13\_single\_subassembly\_qualified\_workcenters}.

Verifier signal: \texttt{demand\_coverage}, \texttt{deadline\_fulfillment}, \texttt{sale\_revenue}, \texttt{supply\_timing\_feasible}, \texttt{po\_delivery\_schedule\_compliance}, \texttt{po\_price\_tier\_compliance}, \texttt{so\_confirmed}, and \texttt{po\_origin\_traceability} fail.

The agent encountered a concrete UI failure while creating the Flint Arena sales order, but kept going without verifying that the customer, product, quantity, price, commitment date, and confirmation state had persisted:

\begin{quote}
``I am noticing that there is a blank line created when the product is clicked, but the customer has not completed their selection. It looks like I need to fill in the current active product''
\end{quote}

\begin{quote}
``It did not seem to select properly; maybe the dropdown closed on its own. Maybe I should press Enter after the active unit price number.''
\end{quote}

Later, the agent made a separate procurement mistake. It recognized that one candidate vendor was late, then still created the late purchase:

\begin{quote}
``I am concerned about the delivery dates \ldots{} Conduit has a late arrival that makes it invalid.''
\end{quote}

\begin{quote}
tool trace: \texttt{\{"type":\ "fill",\ "ref":\ "e90",\ "text":\ "Conduit"\}} \ldots{} \texttt{\{"type":\ "fill",\ "ref":\ "e283",\ "text":\ "84"\}} \ldots{} \texttt{\{"type":\ "fill",\ "ref":\ "e287",\ "text":\ "2103.88"\}}
\end{quote}

What the agent should have done: stop after the Flint Arena form showed a blank or unselected product row, reopen \texttt{S00002}, and verify a confirmed order line for 17 units of \texttt{P43E89D45DC-WTR-CDS-008} at list price with the due date set before creating the remaining orders. For supply, it should have rejected Conduit after identifying the late lead time and selected an on-time route, even if that route required manufacturing or a more expensive finished-goods supplier.

\subsubsection{Computer-Use: Creating Procurement After Proving Stock Was Enough}\label{computer-use-creating-procurement-after-proving-stock-was-enough}

Trial: GPT-5.5, \texttt{2039\_medium\_05\_screened\_buy\_only\_\_3V4SmfP} on \texttt{2039\_medium\_05\_screened\_buy\_only\_all\_seeded}.

Verifier signal: \texttt{new\_spend\_margin\_policy}, \texttt{supply\_timing\_feasible}, \texttt{po\_delivery\_schedule\_compliance}, \texttt{po\_consolidation\_compliance}, \texttt{po\_min\_qty\_compliance}, and \texttt{po\_price\_tier\_compliance} fail.

The agent identified the key fact: only one seeded order should be accepted, and existing finished stock covers it. It then created a PO anyway:

\begin{quote}
``Since there is only one accepted, stock seems to cover it. However, vendor constraints are included. If Cobalt is the only option and stock is not enough, that raises questions. Could I create a zero-quantity PO?''
\end{quote}

\begin{quote}
``I can manually create a PO with the origin S00010. I should consider creating an RFQ or PO from Cipher Solutions for a quantity of 19 at the price of 865.41 \ldots{} This is not strictly necessary, but being explicit is important.''
\end{quote}

The trace then shows the form-entry uncertainty that caused the sourcing-rule failures:

\begin{quote}
``I am noticing an inconsistency with the vendor price; it shows 800.54 instead of 865.41, which is the listed product vendor info.''
\end{quote}

\begin{quote}
``I am noticing the quantity change did not update the amount \ldots{} I should also establish dates and origin.''
\end{quote}

The final answer confirms the mistaken terminal state:

\begin{quote}
``Delivery reserved from finished-goods stock'' and ``Linked traceability PO created/confirmed: P00001''
\end{quote}

What the agent should have done: confirm only the stock-covered \texttt{S00010} order, cancel the other seeded drafts, and stop. Because no new supply was needed for the accepted demand, any nonzero PO introduced avoidable sourcing, timing, and price-tier obligations. If a PO had actually been needed, the agent should have verified the supplier tier price, quantity, expected arrival, and source field from the saved PO form before confirmation.

\section{Reproduction}\label{app:reproduction}

The full task generation pipeline, the released ERP-Bench dataset, and the agent harness adapters used in this paper are available at \url{https://github.com/abhijayrana/erp-bench}. The repository contains the 300 generated tasks under \texttt{tasks/}, the task-generation pipeline under \texttt{erp\_bench/}, the harness adapters under \texttt{agents/}, the Croissant metadata files (\texttt{croissant.json}, \texttt{croissant\_rai.json}), and a \texttt{README.md} with installation, evaluation, and regeneration instructions. The dataset is released under the CC0-1.0 license.

Tasks are evaluated with the open-source \texttt{harbor} runner described in the README. Each task directory follows the layout in Appendix \ref{app:artifact-format}, so a single \texttt{harbor\ run\ -p\ tasks/\textless{}task\_name\textgreater{}\ -a\ \textless{}agent\textgreater{}} invocation builds the Odoo container, runs the agent, and applies the verifier. The README documents reference invocations for the oracle, no-op, and model-backed agents under both Daytona and local Docker environments. Regeneration of the 300-task release from the parametric specification is reproducible from \texttt{erp\_bench/procurement/examples/diverse\_300\_dataset.toml} using the \texttt{generate-tasks} CLI documented in the same README.

\endgroup

\newpage

\end{document}